\documentclass[sigconf]{acmart}

\usepackage{graphicx}
\usepackage{balance}  
\usepackage{graphics}
\usepackage{xspace}
\usepackage[ruled, vlined, linesnumbered]{algorithm2e} 
\usepackage{soul}

\usepackage{amsbsy}
\usepackage{amsthm}
\usepackage{multirow}
\usepackage{mathrsfs}
\usepackage[mathscr]{eucal} 

\usepackage{mathtools}
\usepackage{verbatim}
\usepackage{tabularx}
\usepackage{makecell}
\usepackage{hhline}
\usepackage{placeins}
\usepackage{diagbox}
\usepackage{pifont}

\usepackage{amsfonts}
\usepackage{amsthm}
\usepackage{booktabs}
\usepackage{subcaption}
\usepackage{multirow}
\usepackage{xspace}
\usepackage{enumerate}
\usepackage{subfiles}
\usepackage{soul}
\usepackage{enumitem}
\usepackage{supertabular}

\let\sigproof\proof\let\proof\relax
\let\sigendproof\endproof\let\endproof\relax

\usepackage{amsthm}

\let\proof\sigproof
\let\endproof\sigendproof

\let\oldnl\nl
\newcommand{\nonl}{\renewcommand{\nl}{\let\nl\oldnl}}
\usepackage{dsfont}

\usepackage{array}

\usepackage{enumitem}

\copyrightyear{2023}
\acmYear{2023}
\setcopyright{acmlicensed}
\acmConference[CIKM '23]{Proceedings of the 32nd ACM International Conference on Information and Knowledge Management}{October 21--25, 2023}{Birmingham, United Kingdom}
\acmBooktitle{Proceedings of the 32nd ACM International Conference on Information and Knowledge Management (CIKM '23), October 21--25, 2023, Birmingham, United Kingdom}
\acmPrice{15.00}
\acmDOI{10.1145/3583780.3615038}
\acmISBN{979-8-4007-0124-5/23/10}

\newcommand{\Our}{\textsc{MetaGC}\xspace}
\newcommand{\Ours}{\Our}

\newcommand{\DMoN}{\textsc{DMoN}\xspace}
\newcommand{\MinCutPool}{\textsc{MinCutPool}\xspace}

\newcommand{\DeepWalk}{\textsc{DeepWalk}\xspace}
\newcommand{\DGI}{\textsc{DGI}\xspace}
\newcommand{\GMI}{\textsc{GMI}\xspace}
\newcommand{\Jaccard}{\textsc{GCN-Jaccard}\xspace}
\newcommand{\SVD}{\textsc{GCN-SVD}\xspace}
\newcommand{\ProGNN}{\textsc{ProGNN}\xspace}
\newcommand{\GDC}{\textsc{GDC}\xspace}
\newcommand{\ntov}{\textsc{Node2Vec}\xspace}
\newcommand{\GCC}{\textsc{GCC}\xspace}
\newcommand{\FGC}{\textsc{FGC}\xspace}

\newcommand{\PTDNet}{\textsc{PTDNet}\xspace}
\newcommand{\OurAblOne}{\textsc{MetaGC-X}\xspace}
\newcommand{\OurAblTwo}{\textsc{MetaGC-A}\xspace}

\newcommand{\rom}[1]{\uppercase\expandafter{\romannumeral#1}}
\newcommand\kijung[1]{#1}

\definecolor{Red}{RGB}{244, 124, 124}
\definecolor{Green}{RGB}{162, 222, 147}
\definecolor{Blue}{RGB}{112, 161, 211}

\definecolor{textGreen}{RGB}{186, 255, 169} 

\usepackage{bm}
\newcommand\bolden[1]{{\boldmath\bfseries#1}}

\usepackage{contour}
\usepackage[normalem]{ulem}
\contourlength{0.8pt}

\newcommand{\myuline}[1]{%
  \uline{\phantom{#1}}%
  \llap{\contour{white}{#1}}%
}

\newcommand{\smallsection}[1]{{\noindent {\bolden{\myuline{#1}}}}}
\begin{document}
\title{Robust Graph Clustering via Meta Weighting for Noisy Graphs}


\author{Hyeonsoo Jo}
\affiliation{%
    \institution{Kim Jaechul Graduate School of AI \\ KAIST}
    \city{Seoul}
    \country{Republic of Korea}
}
\email{hsjo@kaist.ac.kr}

\author{Fanchen Bu}
\affiliation{%
    \institution{School of Electrical Engineering \\ KAIST}
    \city{Daejeon}
    \country{Republic of Korea}
}
\email{boqvezen97@kaist.ac.kr}

\author{Kijung Shin}
\affiliation{%
    \institution{Kim Jaechul Graduate School of AI \\ KAIST}
    \city{Seoul}
    \country{Republic of Korea}
}
\email{kijungs@kaist.ac.kr}
\renewcommand{\shortauthors}{Hyeonsoo Jo, Fanchen Bu, \& Kijung Shin}

\begin{abstract}
    \textit{How can we find meaningful clusters in a graph robustly against noise edges?}
    Graph clustering (i.e., dividing nodes into groups of similar ones)
    is a fundamental problem in graph analysis with applications in various fields. Recent studies have demonstrated that graph neural network (GNN) based approaches yield promising results for graph clustering. However, we observe that their performance degenerates significantly on graphs with noise edges, which are prevalent in practice.
    In this work, we propose \Our for robust GNN-based graph clustering. \Our employs a decomposable clustering loss function, which can be rephrased as a sum of losses over node pairs.
    We add a learnable weight to each node pair, and \Our adaptively adjusts the weights of node pairs using meta-weighting so that the weights of meaningful node pairs increase and the weights of less-meaningful ones (e.g., noise edges) decrease. 
    We show empirically that \Our learns weights as intended and consequently outperforms the state-of-the-art GNN-based competitors, even when they are equipped with separate denoising schemes, on five real-world graphs under varying levels of noise. 
    Our code and datasets are available at \url{https://github.com/HyeonsooJo/MetaGC}.
\end{abstract}



\maketitle

\section{Introduction}
\label{sec:intro}
Graphs are a powerful way to represent various systems in physics, bioinformatics, social science, etc.
Real-world graphs usually contain substructures called clusters, where each cluster is a set of similar nodes.
Clusters in real-world graphs have \kijung{useful implications,} 
such as \kijung{social groups} in friendship networks~\cite{girvan2002community,bedi2016community},
functional modules in protein-interaction networks~\cite{chen2006detecting},
and groups of papers on the same topic in citation networks~\cite{sen2008collective}.
Recently,
{graph neural networks (GNNs), which are a class of deep learning models designed to perform inference on \kijung{graph-structured data with} side information (e.g., node attributes),
have received considerable attention.}
GNN-based representation learning 
has shown remarkable performance in various tasks, including node classification, link prediction, and graph classification \cite{kipf2016semi,hamilton2017inductive,zeng2019graphsaint}.

Several GNN-based approaches have been developed also for graph clustering~\cite{tsitsulin2020graph,fettal2022efficient,bianchi2020spectral}. 
In them, GNNs are trained for objectives of graph clustering (e.g., cut and modularity) to produce a (soft) clustering assignment of nodes.
These approaches are effective, especially when abundant node attributes are given, because GNNs are trained end-to-end to exploit both node attributes and graph topology for a considered task.

GNN-based approaches in general are \kijung{known to be} vulnerable to noise edges in graphs, since message passing, the fundamental building block of GNNs, is performed 
through both meaningful edges and noise edges.
Therefore, GNN-based graph-clustering methods also have a common problem of being vulnerable to noise edges.
We observe that their performance degenerates greatly on graphs with noise edges, \kijung{as detailed in the experiment section.}

However, real-world graphs, including social networks \cite{beutel2013copycatch,ferrara2016rise}, auction networks \cite{pandit2007netprobe}, SMS networks \cite{rafique2012graph}, review networks \cite{wang2011review}, computer networks \cite{shin2017densealert}, are often contaminated by noise edges \cite{luo2021learning, dai2022towards} due to click errors \cite{guo2021enhanced}, bots \cite{cai2017detecting, sayyadiharikandeh2020detection}, and spam \cite{li2019spam}, to name a few. 
Recently, several trials have been made on training GNNs to be robust to structural noise.
To alleviate structural noise such as noise edges, some methods~\cite{wu2019adversarial,entezari2020all,jin2020graph,luo2021learning} eliminate noise edges by using similarity of node attributes or some assumptions such as low rank, sparsity, and attribute-smoothness.
Another related line of research has focused on enhancing the robustness of GNNs by modifying the message-passing schemes without explicit graph denoising.
Specifically, in those works, GNNs are designed to be robust for node classification~\cite{feng2021uag,zhang2020gnnguard,zhu2019robust}.
However, methods for improving the robustness of graph clustering have been underexplored.

To address the above problems, we propose \Our (\textbf{\myuline{Meta}}-weighting based \textbf{\myuline{G}}raph \textbf{\myuline{C}}lustering) for robust GNN-based graph clustering against noise edges.
{\Our employs a decomposable clustering loss function, with theoretical justification, and it uses a meta-model to adaptively adjust 
the weight of each pair in the corresponding loss term
(spec., lowering the weights of noise edges).
Both the meta-model and the GNN-based clustering model in \Our are trained end to end for graph clustering. Consequently, \Ours is able to produce better clusters than separately applying graph denoising schemes before graph clustering. 
Moreover, we demonstrate the effectiveness of meta-weighting in comparison to non-meta-weighting-based end-to-end approaches \cite{jin2020graph,luo2021learning}, which we adapt for graph clustering.

Our contributions are listed as follows:
\begin{itemize}[leftmargin=*]
    \item \textbf{Observations}: We show that GNN-based clustering approaches are vulnerable to noise edges.    Theoretically, we 
    define a class of decomposable clustering loss functions (e.g., modularity-based ones)
    and prove that they are suitable for continuous relaxation needed by GNN-based end-to-end learning. 
    \item \textbf{Methodology}: We design \Our for improving the robustness of GNN-based graph clustering. 
    To the best of our knowledge, we are the first {(a) to use meta-weighting for the \textit{robustness} of GNNs and 
    (b) to use meta-weighting specialized in \textit{graph clustering}.}
    \item \textbf{Extensive Experiments}: In our experiments on 5 real-world graphs under 3 levels of noise, we show the advantages of \Our over its state-of-the-art competitors, {even when they   
    use separate
    denoising schemes.}
\end{itemize}

The rest of this paper is organized as follows.
In Sec.~\ref{sec:prelim_related}, we provide some preliminaries and give a brief survey of related work.
In Sec.~\ref{sec:method}, we describe our proposed method.
In Sec.~\ref{sec:exp}, we review our experiments. 
In Sec.~\ref{sec:conclusion}, we conclude our work.


\section{Preliminaries \& Related Work}
\label{sec:prelim_related}
\begin{table}[t]
    \small
    \begin{center}		
    \caption{Frequently-used symbols and definitions.}
    \label{tab:symndef}
    \resizebox{0.9\linewidth}{!}{%
        \begin{tabular}{l|l}
            \toprule 
            \textbf{Symbol}  & \textbf{Definition}\\
            \midrule 
            $G = (V, E)$ & an input graph with nodes $V$ and edges $E$ \\
            $N = |V|$ & the number of nodes \\
            $A \in  \{0,1\}^{N \times N}$ & the adjacency matrix of $G$\\
            $F$ & the dimension of each node's attribute vector \\
            $X \in \mathbb{R}^{N\times F}$ & the node attribute matrix of $G$ \\
            $D = \operatorname{diag}(A\boldsymbol{1}_{N})$ & the degree matrix of $A$ \\
            $d_i = \sum_{i'=1}^{N} A_{ii'}$ & the degree of node $v_i$  \\
            $K$ & the number of clusters \\
            $P \in \mathcal{P} \subseteq [0, 1]^{N \times K}$ & a (soft) cluster assignment matrix \\ 
            \bottomrule
        \end{tabular}
    }
    \end{center}
\end{table}

In this section, we provide some mathematical preliminaries used throughout this paper
and review some related studies.

\subsection{Mathematical Background}
\label{sec:prelim_related:basic}


Let $G = (V, E)$ be an unweighted, undirected,\footnote{For simplicity and due to the nature of the datasets, we focus on unweighted and undirected graphs. Our method is easily extended to weighted and/or directed graphs.} and self-loop-free graph with node set $V= \{v_1,\cdots,v_{|V|}\} $ and edge set $E \subseteq \binom{V}{2}$.
Each edge $(v_i, v_j) = (v_j, v_i) \in E$ {joins} two nodes $v_i\in V$ and $v_j\in V$.
Let $N = |V|$ denote the \textit{number of nodes}. 
Let $A\in \{0,1\}^{N\times N}$ denote the \textit{adjacency matrix} of $G$,
where for two nodes $v_i$ and $v_j$, $A_{ij} = 1$ if and only if $(v_i, v_j) \in E$, i.e., $v_i$ and $v_j$ are joined by an edge.
{The \textit{degree matrix} of $G$ is $D = \operatorname{diag}(A\boldsymbol{1}_{N})$, where each diagonal entry $D_{ii} = d_i = \sum_{i'=1}^{N} A_{ii'}$ is the \textit{degree} of node $v_i$.}

We assume that an attribute vector of dimension $F$ is given for each node,
and we use $X\in \mathbb{R}^{N\times F}$ to denote the corresponding \textit{node attribute matrix},
where the $i$-th row of $X$, denoted by $X_i$, is the node attribute vector of node $v_i$.

Let $K$ be the \textit{number of clusters}.
We call a matrix $P\in [0, 1]^{N \times K}$ a (soft) \textit{cluster assignment matrix} if $\sum_{x = 1}^{K} P_{ix} = 1, \forall 1 \leq i \leq N$, where 
each element
$P_{ix}$ can be interpreted as the probability that we assign node $v_i$ to cluster $x$.
If $P \in \{0, 1\}^{N \times K}$ further holds, then we call $P$ a \textit{deterministic} cluster assignment matrix.
Let $\mathcal{P}$ be the set of all (soft) cluster assignment matrices and $\mathcal{P}^* \subseteq \mathcal{P}$ be the set of all deterministic ones.
We list frequently-used symbols in Table~\ref{tab:symndef}.

\subsection{Graph Clustering \& Quality Functions}

\label{sec:prelim_related:graph_clustering}
Given a graph, the goal of 
\textit{graph clustering} is to divide the nodes into disjoint and exhaustive (i.e., every node is assigned to one group) groups (namely, \textit{clusters}) so that nodes in the same group are more similar to each other than to those in different groups.
Many algorithms have been developed for the problem, and they can be categorized largely into 
partitioning methods~\cite{kernighan1970efficient},
agglomerative methods~\cite{blondel2008fast},
divisive methods~\cite{girvan2002community},
and spectral methods~\cite{newman2013spectral}.

Several measures, including normalized cut~\cite{wu1993optimal} and modularity~\cite{newman2006modularity}, have been used to measure the structural quality (the homogeneity within-cluster nodes and/or the dissimilarity between the cross-cluster nodes) of a given clustering, and they also have been used as objectives for a specific formulation of optimization problems.
Especially, a number of approaches~\cite{blondel2008fast,newman2004fast,lehmann2007deterministic} directly aim to maximize modularity.


\subsection{Meta-weighting}
\label{sec:prelim_related:meta}

Meta-weighting is a method of learning the weights of training samples while minimizing an objective function based on meta-learning.
The weights usually represent the different importance of samples, and they are useful for alleviating class imbalance and reducing the noise in labels. 
The weight of each training sample is obtained by a meta-model, whose parameters are optimized using a small amount of high-quality data without biases and noises along with the learning process of the main model.
Recently, meta-weighting-based schemes outperform traditional rule-based weighting schemes~\cite{ma2018point, cui2019class} in various tasks, including image classification and recommendation~\cite{ren2018learning,shu2019meta,kim2021premere,kim2022meta}.

\smallsection{Our contributions in the context of meta-weighting.} \ To the best of our knowledge, we are first to apply meta-weighting to graph learning. It should be noted, however, that the above techniques are not directly applicable to graph-level tasks (e.g., graph clustering) since 
(a) a graph is not naturally divided into independent samples, and 
(b) which part of a graph is of high quality is typically unknown.
Regarding (a), rather than decomposing data into components, we suggest a novel idea of decomposing a loss function that satisfies the conditions in Definition~\ref{def:dclf}. 
Regarding (b), our study demonstrates that, despite the presence of noise, a meta-model can be effectively trained by using simply distinct batches for it and the clustering model.
That is, we show that a noise-free validation dataset is not mandatory, at least in the context of our problem.


\subsection{GNN-Based Graph Clustering Methods}
\label{sec:prelim_related:gnn_cluster}

Recently, several approaches based on graph neural networks (GNNs) have been proposed for graph clustering~\cite{bianchi2020spectral,tsitsulin2020graph, fettal2022efficient}.
Those approaches are particularly effective when abundant node attributes are given, as GNNs learn to combine node attributes and graph topology for a considered task through end-to-end training.

\MinCutPool~\cite{bianchi2020spectral} uses a graph convolutional network (GCN) followed by a multi-layer perceptron (MLP) and softmax activation.
Its output is a (soft) cluster assignment matrix $P$
(see Sec.~\ref{sec:prelim_related:basic}) that is relaxed to be continuous for end-to-end trainability.
The objective function consists of a continuous relaxation of the normalized cut objective.
\DMoN~\cite{tsitsulin2020graph} employs a similar architecture while it uses a continuous relaxation of modularity (see Sec.~\ref{sec:prelim_related:graph_clustering}) instead of normalized cut.
\GCC~\cite{fettal2022efficient} leverages GCN and the k-means clustering loss \cite{lloyd1982least} to perform node embedding and clustering simultaneously.

\subsection{Graph Denoising and Robust GNNs}
\label{sec:prelim_related:robust}

GNNs in general are vulnerable to noise edges in graphs because message passing, which is the basic operation of GNNs, is performed not only through valid edges but also through noise edges.
Thus, several methods have been proposed to improve the robustness of GNNs by removing potential noise edges from an input graph. 
\Jaccard~\cite{wu2019adversarial} 
{removes an edge if the Jaccard similarity between the node attributes of its two endpoints} is below a predetermined threshold.
\SVD~\cite{entezari2020all} uses a low-rank approximation of the given adjacency matrix instead of the original adjacency matrix.
\ProGNN~\cite{jin2020graph} further assumes the sparsity of the {denoised} adjacency matrix.
{Specifically, in \ProGNN, the denoised adjacency matrix and GNN parameters are learned end to end to minimize jointly 
(a) a classification loss, 
(b) the $\ell_1$ norm (for sparsity) and nuclear norm (for low-rankness) of the adjacency matrix, and 
(c) the difference between attributes of adjacent nodes.
\GDC~\cite{klicpera2019diffusion} creates edges between nodes with high proximity (measured by heat kernel and personalized PageRank) and uses them, and uses such edges, instead of the original edges, for message passing in GNNs.
\PTDNet~\cite{luo2021learning} produces a denoised adjacency matrix through a parameterized denoising network.
The denoised adjacency matrix is subsequently optimized jointly with downstream-task models.
\FGC~\cite{kang2022fine} employs spectral clustering for graph clustering by generating a node-similarity matrix.
This matrix is obtained through a learning process that minimizes a loss function based on both node proximity and filtered attributes,
where the filtered attributes are obtained through a graph Laplacian filter.
All these methods, except for \FGC, yield only a denoised adjacency matrix without providing clustering results.
Therefore, on top of these methods, except for \FGC, we apply \DMoN (see Sec.~\ref{sec:prelim_related:gnn_cluster}) to obtain clustering results so that they can be directly compared with our proposed method for the purpose of clustering (see Sec.~\ref{sec:exp:robust}).

Another related line of research has focused on enhancing the robustness of GNNs by modifying the message-passing schemes without explicit graph denoising.
However, they are designed specifically for node classification~\cite{feng2021uag,zhang2020gnnguard,zhu2019robust}.

\section{Proposed Method: \Ours}
\label{sec:method}
In this section, we introduce our proposed method, \Our (\textbf{\underline{Meta}}-weighting based \textbf{\underline{G}}raph \textbf{\underline{C}}lustering), for robust GNN-based graph clustering.
As shown in Figure~\ref{fig:framework}, \Our consists of a GNN-based clustering model $C$ (see Sec.~\ref{sec:method:cluster})
and a meta-model $M$ (see Def.~\ref{def:dclf} and Sec.~\ref{sec:method:meta}), where $M$ adjusts the weights of node pairs in a decomposable clustering loss function (see Sec.~\ref{sec:method:meta_obj}) that is used to train $C$.
The key idea of \Our is to let $M$ learn to properly adjust the weights of the node pairs (spec., to lower the weights of noise edges) so that the clustering performance of $C$ \kijung{becomes robust.} 

Below, we first introduce the detailed structures of both models ($C$ and $M$).
Then, we discuss the details of the objective function with theoretical analyses.
Lastly, we describe the training process.


\begin{figure}[t!]
    \includegraphics[width=\linewidth]{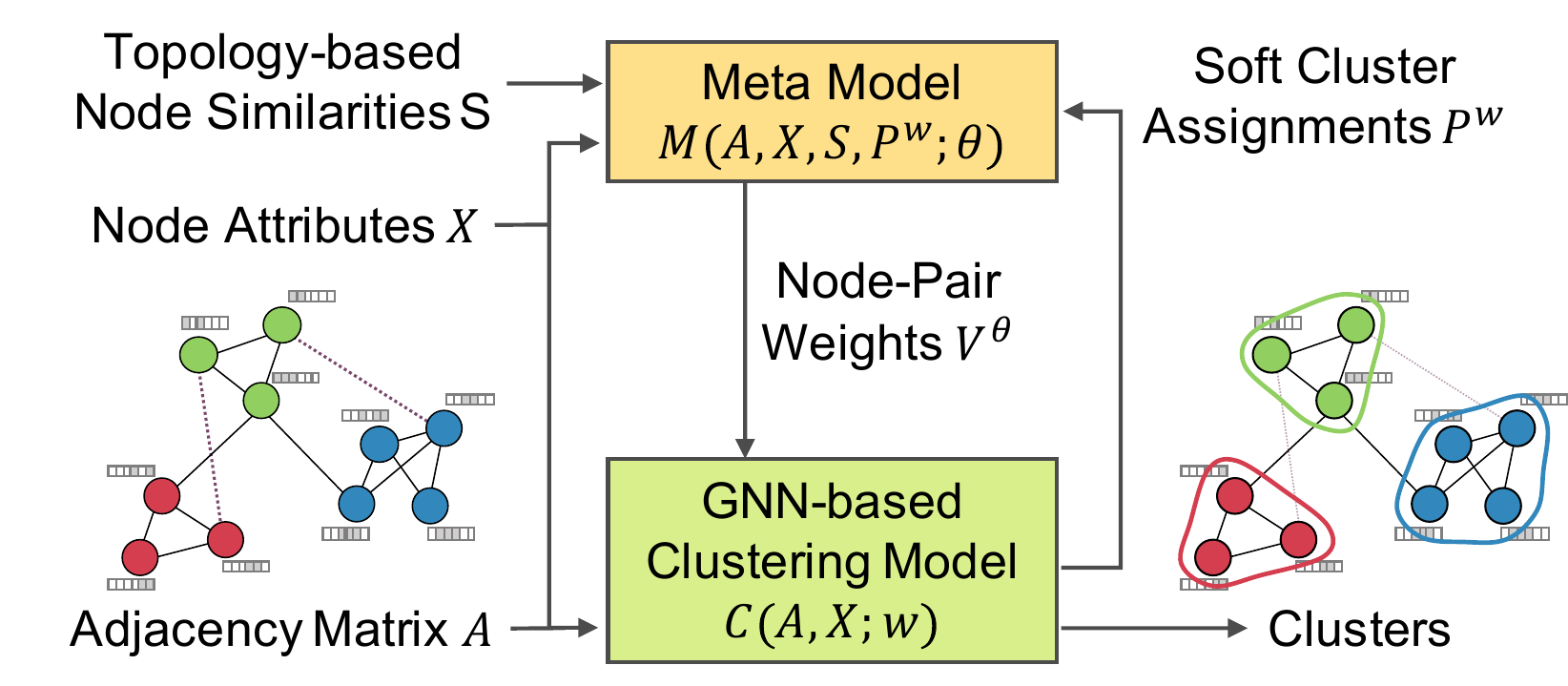}
    \caption{\label{fig:framework} Overall procedure of \Our.
    \Our consists of a GNN-based clustering model $C$ and a meta-model $M$.
    {The meta-model $M$ adjusts the weights of the terms in the modularity-based loss function, which is used to train $C$.}
    } 
\end{figure}




    

\subsection{GNN-based Clustering Model}
\label{sec:method:cluster}

Given an adjacency matrix $A$, a node attribute matrix $X$, and 
the number of clusters $K$,
the clustering model $C$ outputs a (soft) cluster assignment matrix $P^{\boldsymbol{w}}\in[0, 1]^{N \times K}$, where 
$N$ is the number of nodes. 
The clustering model $C$ consists of a GNN, a multilayer perceptron (MLP), and a softmax activation; and $C$ is parameterized by $\boldsymbol{w}$. Specifically,
\begin{equation}
    \hspace{-0.8mm} P^{\boldsymbol{w}} =C(A, X, K; \boldsymbol{w})
    = \operatorname{Softmax}(\operatorname{MLP}(\operatorname{GNN}(A, X))).
    \label{eq:c_model}
\end{equation}
While any GNN models can be used for this purpose, we use a variant with skip connections of GCN~\cite{kipf2016semi}.
Formally, each GNN layer is formulated as the following transformation:
\begin{equation*}
    H^{(i+1)} = \sigma_C(D^{-1/2}AD^{-1/2}H^{(i)}W_{1}+H^{(i)}W_{2}),
\end{equation*}
where $W_1$ and $W_2$ are learnable parameters, $D$ is the degree matrix (see Sec.~\ref{sec:prelim_related:basic}), $\sigma_C$ is an activation function (spec., SELU \cite{klambauer2017self}), 
and $H^{(i)}$ \kijung{for each $i>1$ is the output of the $i$-th GNN layer} and $H^{(0)}= X$.
We use $P^{\boldsymbol{w}}$ to compute a \textit{decomposable} clustering loss function $L^{\boldsymbol{w}}$.
We give the definition of decomposable clustering loss functions first, and the detailed theoretical analyses are deferred to Sec.~\ref{sec:method:meta_obj}.
\begin{definition}[Decomposable clustering loss functions]\label{def:dclf}
    Given $G = (V, E)$ with $|V| = N$, $K \in \mathbb{N}$, and a (soft) cluster assignment matrix $P \in \mathcal{P}$, a clustering loss function $f: \mathcal{P} \to \mathbb{R}$ is \textbf{decomposable}, if there exist constants $c_{ij} = c_{ij}(G), \forall i, j \in [N]$ s.t.
    \begin{equation}\label{eq:decomposable}
        f(P) = \sum_{i = 1}^N \sum_{j = 1}^N c_{ij} P_i \cdot P_j, 
    \end{equation}
    where $P_i$ denotes the $i$-th row of $P$.\footnote{WLOG, we assume $f$ does not contain constant terms.}
\end{definition}

\subsection{Meta-Model}\label{sec:method:meta}
The inputs of the meta-model $M$ are
(a) the adjacency matrix $A$, 
(b) the node attribute matrix $X$, 
(c) the topology-based node-similarity matrix $S$ (spec., we use the Adamic-Adar indices~\cite{adamic2003friends}), and 
(d) the soft cluster assignment matrix $P^{\boldsymbol{w}}$ (i.e., the output of $C$);
and $M$ outputs the node-pair weight matrix $V^{\boldsymbol{\theta}}\in \mathbb{R}_{>0}^{N\times N}$ for weighting the terms in the final loss function (i.e., Eq.~\eqref{eq:overall_loss} in Sec.~\ref{sec:method:meta_obj}).
The parameters of $M$ are denoted by $\boldsymbol{\theta}$.
Specifically,
\begin{align}
    V^{\boldsymbol{\theta}} & = M(A, X, S, P^{\boldsymbol{w}};\boldsymbol{\theta}) \nonumber \\ & = \sum\nolimits_{r=1}^{R} \alpha^{(r)} \cdot \sigma_M \left( (Z^{(r)}(Z^{(r)})^{T}) \odot Y^{(r)}\right),
    \label{eq:m_model}
\end{align}
where $\odot$ denotes the element-wise product, 
$\sigma_M$ is an activation function (spec., we use sigmoid),
and $R$ is the number of the features of node pairs.
For each $1 \leq r \leq R$, $Y^{(r)}\in\mathbb{R}^{N\times N}$ is a feature matrix for node pairs,
the scalar $\alpha^{(r)}$ is the learnable weight\footnote{The weights are normalized so that $\sum_{r=1}^{R} \alpha^{(r)} = 1$ holds.} for $Y^{(r)}$, and 
the symmetric matrix $Z^{(r)}(Z^{(r)})^{T}\in\mathbb{R}^{N\times N}$ with $Z^{(r)}=\operatorname{MLP}^{(r)}(X)$ is used as the attention matrix for $Y^{(r)}$.
While any features of node pairs can be employed, after a preliminary study, we use the following three features (i.e., $R = 3$) for each node pair $(v_i,v_j)$:
(a) if $A_{ij} = 0$, $(Y^{(1)}_{ij}, Y^{(2)}_{ij}, Y^{(3)}_{ij}) = (1, 0, 0)$;
(b) if $A_{ij} = 1$, $(Y^{(1)}_{ij}, Y^{(2)}_{ij}, Y^{(3)}_{ij}) = (1, S_{ij}, L^{\boldsymbol{w}}_{ij})$,
where $S$ is the topology-based node-similarity matrix,
and $L^{\boldsymbol{w}}_{ij}$ is a decomposable clustering loss function computed using $P^{\boldsymbol{w}}$ (i.e., the output of $C$), whose details with theoretical analyses will be provided in Sec.~\ref{sec:method:meta_obj}.



\subsection{Learning Objective}
\label{sec:method:meta_obj}

The objective function of \Our includes a \textit{decomposable} clustering loss function (see Def.~\ref{def:dclf}).
We shall show that decomposable clustering loss functions are \textit{expectation-conforming}, which is a desirable property for continuous relaxation needed by GNN-based end-to-end learning.
Given a soft clustering assignment matrix $P$, we can interpret each entry $P_{ix}$ as the probability that we assign node $v_i$ to the cluster $x$.
Thus, we may see $P$ as a random variable taking values in $\mathcal{P}^*$ \kijung{(see Sec.~\ref{sec:prelim_related:basic})} and the probability that $P$ corresponds to a deterministic $P^* \in \mathcal{P}^*$ is
\begin{equation}\label{eq:prob_soft_hard}
\mathbb{P}(P = P^*) = \prod\nolimits_{i = 1}^{n} P_{i, S(i;P^*)},
\end{equation}
where $S(i;P^*)$ is the cluster to which $P^*$ assigns $v_i$ (i.e., $P^*_{ix} = 1$ if and only if $x = S(i;P^*)$).
Using such a perspective, we first give the formal definitions of expectation-conforming and decomposable clustering loss functions.
\begin{definition}[Expectation-conforming clustering loss functions]\label{def:ecclf}
Given $G = (V, E)$ with $|V| = N$, $K \in \mathbb{N}$, and a (soft) cluster assignment matrix $P \in \mathcal{P}$,
a clustering loss function $f: \mathcal{P} \to \mathbb{R}$ is \textbf{expectation-conforming}, if
\begin{equation}\label{eq:exp_conf}
    f(P) = \sum\nolimits_{P^* \in \mathcal{P}^*} \mathbb{P}(P = P^*) f(P^*), \forall P \in \mathcal{P}.
\end{equation}
\end{definition}
The intuition is that, if $f$ satisfies Eq.~\eqref{eq:exp_conf}, then for each (soft) cluster assignment matrix $P$, the value of $f$ directly taken on $P$ is equal to the \textit{expectation} of the value of $f$ when we see $P$ as a random variable taking values in $\mathcal{P}^*$.



\begin{lemma}\label{lem:min_exp_conf}
    If $f: \mathcal{P} \to \mathbb{R}$ is expectation-conforming, then
    (1) $\min_{P \in \mathcal{P}} f(P) = \min_{P^* \in \mathcal{P}^*}$ $f(P^*)$, and
    (2) for each $P \in \arg\min_{P \in \mathcal{P}}$ $f(P)$,
    $\mathbb{P}(P = P^*) > 0 \Rightarrow P^* \in \arg\min_{P^* \in \mathcal{P}^*} f(P^*)$.
\end{lemma}
\begin{proof}
    Since $\mathcal{P}^* \subseteq \mathcal{P}$, it is trivial that $\min_{P \in \mathcal{P}} f(P) \leq \min_{P^* \in \mathcal{P}^*}$ $f(P^*)$.
    On the other hand, since for each $P \in \mathcal{P}$, $\sum_{P^* \in \mathcal{P}^*} \mathbb{P}(P = P^*) = 1$,
    $f(P) = \sum_{P^* \in \mathcal{P}^*} \mathbb{P}(P = P^*) f(P^*) \geq \min_{P^* \in \mathcal{P}^*} f(P^*)$, completing the proof for (1).

    Regarding (2), suppose the opposite, i.e., $\exists P \in \arg\min_{P \in \mathcal{P}} f(P),$ $P^*_0 \in \mathcal{P}^* \notin \arg\min_{P^* \in \mathcal{P}^*} f(P^*), \mathbb{P}(P = P^*) > 0$, then it is easy to see that
    $f(P) = \sum_{P^* \in \mathcal{P}^*} \mathbb{P}(P = P^*) f(P^*) > \min_{P^* \in \mathcal{P}^*} f(P^*)$, which, combined with (1), completes the proof.
\end{proof}
\smallsection{Why is being expectation-conforming desirable?} 
{When we use an expectation-conforming clustering loss function $f$, if we reach a global minimum after training, then it is guaranteed that we can obtain an optimal solution to the graph clustering problem (i.e., an optimal deterministic clustering assignment) from the trained (soft) clustering assignment matrix.} 



\begin{lemma}\label{lem:decomposable_then_exp_conf}
    If $f: \mathcal{P} \to \mathbb{R}$ is decomposable, then $f$ is expectation-conforming.
\end{lemma}
\begin{proof}
    By Def.~\ref{def:dclf}, Eq.~\ref{eq:prob_soft_hard}, and 
    linearity of expectation,
    it suffices to show that each term of $f$ is expectation-conforming.
    For the $P_i \cdot P_j$ terms,
    $\sum_{P^* \in \mathcal{P}^*} \mathbb{P}(P = P^*) P_i^* \cdot P_j^* 
    = \sum_{P_i^*, P_j^*} \mathbb{P}(P_i = P_i^*) \mathbb{P}(P_j = P_j^*) P_i^* \cdot P_j^* 
    = \sum_{x} \mathbb{P}(P_i = P_i^*, P_{ix}^* = 1) \mathbb{P}(P_j = P_j^*, P_{jx}^* = 1)
    = \sum_{x} P_{ix} P_{jx}
    = P_i \cdot P_j$,
    where we have used the independence between $P_i$ and $P_j$, for all $i \neq j$,
    completing the proof.
\end{proof}

Specifically, in \Our, we use a modularity-based loss function (Eq.~\eqref{eq:modularity_loss}).\footnote{
Modularity measures how much a cluster's edge count ($A_{ij}$'s) exceeds the expected edge count ($d_i d_j / 2|E|$'s) under a null model preserving the degree distribution.
}
As mentioned before, modularity~\cite{newman2006modularity} is a representative objective for clustering.
We use the output $P^{\boldsymbol{w}}$ of the clustering model $C$ (see Sec.~\ref{sec:method:cluster}) to compute the loss function.
Formally,
\begin{equation}\label{eq:modularity_loss}
     \tilde{Q}(\boldsymbol{w}) = \sum\nolimits_{i, j = 1}^N \underbrace{\frac{1}{2|E|}(A_{ij}-\frac{d_{i}d_{j}}{2|E|}) P^{\boldsymbol{w}}_{i} \cdot P^{\boldsymbol{w}}_{j}}_{=\tilde{Q}_{ij}(\boldsymbol{w})},
\end{equation}
where we have decomposed modularity over node pairs.
Note that, $\tilde{Q}(\boldsymbol{w})$ is one specific example of decomposable clustering loss functions, which are denoted above by $L^{\boldsymbol{w}}$.

\begin{lemma}\label{lem:modularity_is_decomposible}
    The function $\tilde{Q}$ is decomposable, and thus it is expectation-conforming.
\end{lemma}
\begin{proof}[Proof of Lem.~\ref{lem:modularity_is_decomposible}]
    $\Tilde{Q}$ is decomposable with
    $c_{ij} = \frac{1}{2|E|}(A_{ij} - \frac{d_i d_j}{2|E|}), \forall i, j \in [N]$.
\end{proof}
\smallsection{Remark.} Due to the NP-hardness of modularity optimization~\cite{brandes2006maximizing}, finding a globally optimal solution is non-trivial in general, and we are not claiming that using an expectation-conforming function makes it easier to reach an optimal solution. Instead, we are claiming that applying a continuous extension to an expectation-conforming function does not introduce any ``bad'' minima~\cite{karalias2021neural} (minima of the continuous extension but not the original problem).


In contrast, the normalized cut~\cite{wu1993optimal,bianchi2020spectral} is not decomposable nor expectation-conforming in general.

\begin{lemma}\label{lem:ncut_not_exp_conf}
   In general, $\operatorname{NC} = \operatorname{NC}(P) = -\operatorname{Tr}(P^T \Tilde{A} P) / \operatorname{Tr}(P^T \Tilde{D} P)$ is not expectation-conforming, and thus it is not decomposable either.
\end{lemma}
\begin{proof}
    We expand $\operatorname{NC}$ as $\operatorname{NC}(P) = -\frac{\sum_{i, j}(P_i \cdot P_j)\tilde{A}_{ij}}{\sum_{i} (P_i \cdot P_i)\tilde{D}_ii}$.
    Consider a simple case where $G$ consists of two nodes and one edge between them with $K = 2$.
    We have $\operatorname{NC}(P) = -\frac{2 P_1 \cdot P_2}{P_1 \cdot P_1 + P_2 \cdot P_2} \neq
    \sum_{P^* \in \mathcal{P}^*} \mathbb{P}(P = P^*) \operatorname{NC}(P^*) = P_{11} P_{21} \operatorname{NC}((\begin{smallmatrix}1 & 0\\ 1 & 0\end{smallmatrix})) + P_{12} P_{22} \operatorname{NC}((\begin{smallmatrix}0 & 1\\ 0 & 1\end{smallmatrix})) = -P_{11} P_{21} - P_{12} P_{22} = -P_1 \cdot P_2$.
    The counterexamples when $G$ has more nodes or edges can be constructed similarly.
\end{proof}

In order to integrate the meta-weighting strategy into our decomposable clustering loss function, 
the loss at each node pair $v_i$ and $v_j$ is weighted by $V^{\boldsymbol{\theta}}_{ij}\in \mathbb{R}_{>0}$ produced by the meta-model $M$ (see Sec.~\ref{sec:method:meta}).
Formally, the final loss function is
\begin{equation}
   L(\boldsymbol{w}, \boldsymbol{\theta}) = \left(\sum\nolimits_{i, j=1}^{N} V^{\boldsymbol{\theta}}_{ij}L^{\boldsymbol{w}}_{ij}\right)  + \lambda R^{\boldsymbol{w}},
    \label{eq:overall_loss}
\end{equation}
where 
$\lambda$ denotes the regularization rate, and $R^{\boldsymbol{w}}=\frac{\sqrt{K}}{N}|| \sum_{i=1}^{N} P^{\boldsymbol{w}}_{i}||_{F}-1$ is a collapse regularization term introduced in~\cite{tsitsulin2020graph}.
For ease of explanation, we also define the loss at each node as follows:
\begin{equation}
L_{i}(\boldsymbol{w}, \boldsymbol{\theta}) = \left(\sum\nolimits_{j=1}^{N} V^{\boldsymbol{\theta}}_{ij}L^{\boldsymbol{w}}_{ij}\right) + \frac{\lambda}{N} R^{\boldsymbol{w}}.
\label{eq:node_loss}
\end{equation}

\begin{algorithm}[t!]
    \small
    \caption{Learning Algorithm of \Our}
    \label{alg:metagc}
    \SetAlgoLined
    \SetKwInOut{Input}{input}
    \SetKwInOut{Output}{output}
    \DontPrintSemicolon
    \Input{
    (a) graph $G = (V, E)$, 
    (b) node attribute matrix $X$, \linebreak
    (c) number of clusters $K$, 
    (d) number of epochs $T$, \linebreak
    (e) number of batches $b$}    
    \Output{clustering model parameter $\boldsymbol{w}$}        
    Initialize clustering/meta-model parameter $\boldsymbol{w}$/$\boldsymbol{\theta}$ \\    
    \For  {$epoch=1$ to $T$} {
        \For {$itr=1$ to $N/b$} {
            $B_{C}, B_{M} \gets$ two disjoint subsets of $V$ of size $b$ sampled uniformly at random\\
            \tcp{Meta-model update}
            $\boldsymbol{w}' \leftarrow \boldsymbol{w} - \eta\sum\nolimits_{v_i \in B_{C}}\nabla_{\boldsymbol{w}}L_{i}(\boldsymbol{w}, \boldsymbol{\theta})$ \\
            $\boldsymbol{\theta} \leftarrow \boldsymbol{\theta} - \mu\sum\nolimits_{v_i \in B_{M}}\nabla_{\boldsymbol{\theta}}\tilde{Q}_{i}(\boldsymbol{w}')$ \\
            \tcp{Clustering model update}
            $\boldsymbol{w} \leftarrow \boldsymbol{w} - \eta\sum\nolimits_{v_i \in B_{C}}\nabla_{\boldsymbol{w}}L_{i}(\boldsymbol{w}, \boldsymbol{\theta})$
        }
    } 
    \Return{$\boldsymbol{w}$}    
\end{algorithm}

\subsection{Overall Training Procedure}
\label{sec:method:overall}

In Alg.~\ref{alg:metagc}, we provide the pseudocode of the training process, where we {alternatingly} optimize the meta-model and the clustering model, following the general procedure of meta-weighting~\cite{ren2018learning,shu2019meta}.

While the general procedure requires noise-free validation data, it is typically unknown which part of the input graph is noise-free.
Thus, without such requirements, \Our employs distinct batches (i.e., subsets) of nodes and their corresponding incident edges for training the meta-model and the clustering model. This approach is founded on the belief that the input graph contains sufficient information to effectively train the meta-model, even in the presence of noise edges, and this belief is validated in Sec.~\ref{sec:exp:robust}.

\smallsection{Meta-model update.}
In each training step $t$, in order to update the parameters $\boldsymbol{\theta}_{t}$ of the meta-model $M$, 
the parameters $\boldsymbol{w}_{t}$ of clustering model $C$ are employed.
We first update $\boldsymbol{w}_{t}$ once using the weighted loss function (i.e., Eq.~\eqref{eq:overall_loss}) using a batch $C_{t}$ of nodes as follows:
\begin{equation}
  \boldsymbol{w}'_t \leftarrow \boldsymbol{w}_{t} - \eta\sum\nolimits_{v_i \in C_{t}}\nabla_{\boldsymbol{w}}L_{i}(\boldsymbol{w}, \boldsymbol{\theta}_{t}) \big|_{\boldsymbol{w}=\boldsymbol{w}_{t}},
    \label{eq:meta_update_1}
\end{equation}
where $\eta$ is the learning rate for $C$.


{Using the updated parameters $\boldsymbol{w}'$, we update the parameters $\boldsymbol{\theta}_{t}$ of the meta-model using another batch $M_{t}$ of nodes and the unweighted loss function:\footnote{{Note that the weighted loss function \eqref{eq:overall_loss} cannot be used to update $\boldsymbol{\theta}$ since it is trivially minimized when each $V^{\boldsymbol{\theta}}_{ij}=0$.
}}
\begin{equation}
  \boldsymbol{\theta}_{t+1} \leftarrow \boldsymbol{\theta}_{t} - \mu\sum\nolimits_{v_i \in M_{t}}\nabla_{\boldsymbol{\theta}}\tilde{Q}_{i}(\boldsymbol{w}'_t) \big|_{\boldsymbol{\theta}=\boldsymbol{\theta}_{t}},
    \label{eq:meta_update_2}
\end{equation}
where $\mu$ is $M$'s learning rate, and $\tilde{Q}_{i} = \sum_j \Tilde{Q}_{ij}$ (see Eq.~\eqref{eq:modularity_loss}).

\smallsection{Clustering model update.}
{In each training step $t$, the clustering model $C$ is updated after the meta-model update.
The parameters $\boldsymbol{w}_{t}$ of $C$ are updated using the weighted loss function (i.e., Eq.~\eqref{eq:overall_loss}) with a batch $C_{t}$ of nodes and the meta-model $M$ with its updated parameters $\boldsymbol{\theta}_{t+1}$ as follows:
\begin{equation}
   \boldsymbol{w}_{t+1} \leftarrow \boldsymbol{w}_{t} - \eta\sum\nolimits_{v_i \in C_{t}}\nabla_{\boldsymbol{w}}L_{i}(\boldsymbol{w}, \boldsymbol{\theta}_{t+1})\big|_{\boldsymbol{w}=\boldsymbol{w}_{t}}.
    \label{eq:cluster_update}
\end{equation}}


\section{Experiments}
\label{sec:exp}
In this section, we evaluate \Our to answer the Q1-Q3:
\begin{enumerate}[leftmargin=*]
	\item[Q1.] \textbf{Robustness \& Accuracy}: Is \Our more robust and accurate than the competitors on noisy graphs?
	\item[Q2.] \textbf{Effectiveness of Meta-Weighting}: Does the meta-model in \Our properly adjust the weights of loss terms? 
	\item[Q3.] \textbf{Ablation Study}: Does \kijung{each component} of \Our contribute to performance improvement?
\end{enumerate}

\subsection{Experiment Settings}
\label{sec:exp:settings}
\smallsection{Hardware.} We run all experiments on a workstation with an Intel Xeon 4214 CPU, 512GB RAM, and RTX 8000 GPUs.

\begin{table}[t!]
	\begin{center}
		\caption{\label{tab:dataset} Summary of the real-world datasets}
		\small
			\begin{tabular}{r|r|r|r|r}
				\toprule 
				\textbf{Name} & \textbf{\# Nodes} & \textbf{\# Edges} & \textbf{\# Attributes} & \textbf{\# Classes}\\
				\midrule
				Cora & 2,708 & 5,278 & 1,433 & 7 \\
				Cora-ML & 2,995 & 8,158 & 2,879 & 7 \\
			    Citeseer & 3,327 & 4,552 & 3,703 & 6 \\
			    Amazon-Photo & 7,535 & 119,081 & 745 & 8 \\
				Pubmed & 19,717 & 44,324 & 500 & 3 \\
				\bottomrule
			\end{tabular}
	\end{center}
\end{table}

\smallsection{Datasets.} 
We use 5 real-world datasets: 
4 citation graphs (Cora, Cora-ML, Citeseer, Pubmed ~\cite{sen2008collective}) and 
a co-purchase graph (Amazon-Photo~\cite{shchur2018pitfalls}). 
For all graphs, self-loops are ignored. 
Some basic statistics of the graphs are provided in Table~\ref{tab:dataset}.

\smallsection{Noisy-graph Generation.} 
For each dataset, we generate noise edges that do not exist in the original graph and add them to the graph.
Specifically, the noise edges are chosen uniformly at random among those whose endpoints belong to different classes.
Noise levels \rom{1}, \rom{2} and \rom{3} indicate that the ratio between the number of noise edges and that of the existing edges are 30\%, 60\%, and 90\%, respectively.
For each dataset and each noisy level, we generate five noisy graphs, and all experimental results are averaged over them.

\smallsection{Competitors.} 
We compare \Our with 13 competitors that are divided into three categories:
(a) four \textit{node-embedding-based} methods (\DeepWalk~\cite{perozzi2014deepwalk}, 
\ntov~\cite{grover2016node2vec}, 
\DGI~\cite{velivckovic2018deep}, 
and \GMI~\cite{peng2020graph}), whose outputs are clustered by K-means++~\cite{vassilvitskii2006k},
(b) three \textit{GNN-based graph clustering} methods (\MinCutPool~\cite{bianchi2020spectral}, 
\DMoN~\cite{tsitsulin2020graph}, and \GCC~\cite{fettal2022efficient})
(see Sec.~\ref{sec:prelim_related:gnn_cluster} for the details), and 
(c) six \textit{graph denoising} methods (\Jaccard~\cite{wu2019adversarial}, 
\SVD~\cite{entezari2020all}, 
\ProGNN~\cite{jin2020graph},
\GDC~\cite{klicpera2019diffusion},
\PTDNet~\cite{luo2021learning},
and \FGC~\cite{kang2022fine}).\footnote{Technically speaking, \GDC and \FGC are graph augmentation methods.}
All the \textit{graph denoising} methods, except for \FGC, are combined with \DMoN for graph clustering. Note that \DMoN aims to maximize a continuous relaxation of modularity, which is used as an evaluation metric.

Some modifications are needed for the graph denoising methods so that they can be used for graph clustering.
For \SVD and \GDC, which generate weighted graphs, we have observed that
applying \DMoN directly on weighted graphs 
impairs the performance of graph clustering.
Therefore, we convert each generated weighted graph to an unweighted one by selecting the $|E|$ edges with the largest weights, where $|E|$ is the number of edges in the original graph.
For \ProGNN and \PTDNet (see Sec.~\ref{sec:prelim_related:robust} for more details),
we replace their classification loss with our clustering loss (see Eq.~\eqref{eq:modularity_loss}).

For \Ours and all its competitors, we use the ground-truth number of classes in each graph as the target number of clusters $K$. See Apppendix~\ref{appendix:param_settings} for more hyperparameter settings.

\smallsection{Evaluation Metrics.}
To evaluate \Our and the competitors, we use both topology- and correlation-based metrics.
As a topology-based metric, we use the modularity 
in the original graph without injected noises.
As correlation-based metrics, we use the pairwise F1 Score (F1 Score) and the normalized mutual information (NMI) between the given classes of nodes and the cluster assignments of nodes.
For each output (soft) cluster assignment matrix of the methods, we apply each of the evaluation metrics after converting it to a deterministic one by setting the maximum value of each row to 1 and the rest to 0. 
Notably, using each of the metrics alone, there exist cases where a higher value may not necessarily mean more meaningful clustering~\cite{liu2019evaluation,lancichinetti2011limits,chicco2020advantages}. 
\kijung{We also observe some specific cases} in our experiments (see Sec.~\ref{sec:exp:robust}).
Therefore, it is important to take all of them into consideration.

On each noisy graph, three independent trials are conducted, and thus for each dataset and each noise level, we report the average results over the 15 trials (5 noisy graphs $\times$ 3 independent trials).

\subsection{Q1. Robustness \& Accuracy}
\label{sec:exp:robust}
In Table~\ref{tab:result}, for each dataset, each noise level, each method, and each evaluation metric, we report the mean and standard deviation of the results over 15 trials. 

For each dataset and each method, we also report the average rank (AR) over all the evaluation metrics. 
With regard to AR, tests of statistical significance are performed between the proposed method \Our and each competitor over 15 independent random trials.
Specifically, for each competitor, the null hypothesis is that there is no significant superiority of \Our over the competitor w.r.t AR.
One-tailed $t$-tests are employed to ascertain whether the AR of \Our is significantly better than that of each competitor. 
The results of the tests are reported as follows: 
* means $p < 0.05$, 
** means $p < 0.01$, 
and *** means $p < 0.001$.

First of all, \Our performs best overall and ranks first in every dataset w.r.t the average rank.
Specifically, \Our achieves an average rank of $1.2$ to $3.3$ among all the $14$ considered methods on the five datasets.
Below, we discuss the results in detail. 

Since \DeepWalk and \ntov use only graph topology without node attributes, their performance w.r.t modularity, which is a topology-based metric, is favorable. 
However, they are much less competitive than \Our w.r.t F1 Score and NMI, which are correlated-based metrics. 

In some cases, \DGI and \GMI perform better than \Our w.r.t NMI,
but in most cases, they perform significantly worse w.r.t the other two metrics, especially F1 Score.
Moreover, their performance decreases on large graphs (Amazon-Photo and Pubmed).

Compared to the two GNN-based graph clustering methods \textit{without meta-weighting} (\MinCutPool, \DMoN, and \GCC), the performance superiority of \Our 
becomes more significant as the noise level increases, showing that meta-weighting indeed enhances the robustness
(see Sec.~\ref{sec:exp:effective} for more discussions on the effectiveness of meta-weighting).
In some cases, \MinCutPool and \GCC output only one cluster that contains all nodes, which results in a high yet meaningless F1 Score (see the low NMI and modularity).

Recall that each of the graph denoising methods except \FGC is actually \DMoN with a separate or combined denoising process (see Sec.~\ref{sec:exp:settings}), and \DMoN aims to maximize a continuous relaxation of modularity, which is used as an evaluation metric.
\Our outperforms all of them w.r.t the average rank on every dataset.
Notably, even though \SVD and \GDC use the ground-truth number of the edges in the original noise-free graph, which is unknown to \Our, they perform consistently worse than \Our in all respects.
\Jaccard sometimes shows slightly better performance than \Our w.r.t F1 Score, but \Our performs better w.r.t the other two metrics in almost all the other cases (except for the performance w.r.t modularity on Pubmed in noise level \rom{1}).
\ProGNN and \PTDNet both have a clear drawback compared to \Our.
Moreover, \ProGNN is not scalable for large graphs due to its $O(N^{3})$ time complexity.
\PTDNet is not scalable for large graphs either due to its space complexity associated with its parameterized denoising network, which generates a denoised adjacency matrix.

\subsection{Q2. Effectiveness of Meta-Weighting}
\label{sec:exp:effective}
We shall show that the meta-model in \Our effectively distinguishes real edges and noise edges by assigning high weights to real edges and low weights to noise edges.
We see this as a binary classification task,
and in Table~\ref{tab:effective}, for each dataset and each noise level, we report
the PRAUC (the area under the Precision-Recall curve) 
and 
HITS@10\% (i.e., the proportion of real edges among the top-10\% edges with highest weights) value.
We also include a baseline representing the expected value of PRAUC and HITS@10\% when we randomly assign the weights, which is equal to the proportion of real edges among all the edges.
We can see that both PRAUC and HITS@10\% of \Our are consistently higher than the expected value.
Moreover, HITS@10\% values are consistently close to $1$, which means that in all the cases, almost all top-10\% edges with the highest weights assigned by the meta-model are real edges, as intended.
In Figure~\ref{fig:prauc}, we present the detailed Precision-Recall curves at different noise levels for each dataset.

\begin{table*}[h!]
    \centering
    \caption{\label{tab:result}
    (Q1) \underline{\smash{Robustness \& Accuracy.}} 
    \Our shows the best overall performance.
    It ranks the first in all datasets w.r.t the average rank (AR).
    Statistical significance of the AR differences between \Our and each of its competitors is reported  at the following levels: *p < 0.05, **p < 0.01, ***p < 0.001.
O.O.T.: out of time ($>$ 6 hours). O.O.M: out of (GPU) memory.
    For each setting (each column), the best, second-best and third-best results are in \colorbox{Red!40}{red}, \colorbox{Blue!40}{blue}, and \colorbox{Green!40}{green}, respectively.}
    \begin{subtable}{\textwidth}
        \vspace{1mm}
        \centering
        \resizebox{\textwidth}{!}{
            \begin{tabular}{ c||c|c|c||c|c|c||c|c|c||c } 
                \hline
                 \textbf{Noise Level} & \multicolumn{3}{c||}{\rom{1}} & \multicolumn{3}{c||}{\rom{2}} & \multicolumn{3}{c||}{\rom{3}} & \multirow{2}{*}{\textbf{{AR}}} \\
                 \cline{1-10}
                 \textbf{Metric} & F1 Score &  NMI & Modularity &F1 Score &  NMI & Modularity & F1 Score &  NMI & Modularity & \\
                \hline
                \hline
\DeepWalk & 0.405$\pm$0.048 & 0.465$\pm$0.008 & \colorbox{Blue!40}{0.689$\pm$0.006} & 0.297$\pm$0.022 & 0.389$\pm$0.010 & \colorbox{Green!40}{0.659$\pm$0.007} & 0.256$\pm$0.014 & 0.352$\pm$0.012 & \colorbox{Green!40}{0.641$\pm$0.007} & {${6.0}^{***}$}\\
\ntov & 0.410$\pm$0.043 & 0.464$\pm$0.006 & \colorbox{Red!40}{0.690$\pm$0.005} & 0.296$\pm$0.023 & 0.389$\pm$0.008 & \colorbox{Blue!40}{0.660$\pm$0.004} & 0.261$\pm$0.017 & 0.359$\pm$0.011 & \colorbox{Blue!40}{0.642$\pm$0.007} & {${5.4}^{***}$}\\
\DGI & 0.230$\pm$0.010 & 0.287$\pm$0.003 & 0.151$\pm$0.007 &0.198$\pm$0.009 & 0.239$\pm$0.004 & 0.141$\pm$0.010 &0.183$\pm$0.006 & 0.203$\pm$0.013 & 0.122$\pm$0.013 &{${9.7}^{***}$}\\
\GMI & 0.099$\pm$0.004 & 0.021$\pm$0.001 & -0.003$\pm$0.001 &0.103$\pm$0.005 & 0.025$\pm$0.001 & -0.002$\pm$0.001 &0.109$\pm$0.006 & 0.030$\pm$0.001 & -0.002$\pm$0.001 &{${11.7}^{***}$}\\
\hline
\hline
\MinCutPool & 0.464$\pm$0.000 & 0.000$\pm$0.000 & 0.000$\pm$0.000 &0.464$\pm$0.000 & 0.000$\pm$0.000 & 0.000$\pm$0.000 &0.464$\pm$0.000 & 0.000$\pm$0.000 & 0.000$\pm$0.000 &{${9.4}^{***}$}\\
\DMoN & \colorbox{Green!40}{0.556$\pm$0.049} &  \colorbox{Blue!40}{0.533$\pm$0.041} &  0.609$\pm$0.036 &\colorbox{Red!40}{0.528$\pm$0.028} &  \colorbox{Blue!40}{0.494$\pm$0.025} &  0.599$\pm$0.023 &\colorbox{Green!40}{0.470$\pm$0.033} &  \colorbox{Green!40}{0.425$\pm$0.036} &  0.531$\pm$0.050 &{${3.3}^{***}$}\\
\GCC & 0.538$\pm$0.022 & 0.501$\pm$0.039 & 0.619$\pm$0.034 &0.469$\pm$0.007 & 0.377$\pm$0.019 & 0.540$\pm$0.027 &0.459$\pm$0.006 & 0.353$\pm$0.018 & 0.526$\pm$0.024 &{${5.9}^{***}$}\\
\hline
\hline
\Jaccard & \colorbox{Blue!40}{0.557$\pm$0.049} &  \colorbox{Blue!40}{0.533$\pm$0.040} &  0.610$\pm$0.036 &\colorbox{Blue!40}{0.525$\pm$0.034} &  \colorbox{Green!40}{0.493$\pm$0.028} &  0.597$\pm$0.032 &\colorbox{Blue!40}{0.473$\pm$0.034} &  \colorbox{Blue!40}{0.431$\pm$0.038} &  0.538$\pm$0.052 &{${3.1}^{***}$}\\
\SVD & 0.390$\pm$0.004 & 0.365$\pm$0.009 & 0.497$\pm$0.002 &0.408$\pm$0.005 & 0.379$\pm$0.004 & 0.506$\pm$0.006 &0.403$\pm$0.005 & 0.374$\pm$0.017 & 0.507$\pm$0.011 &{${7.4}^{***}$}\\
\GDC & 0.514$\pm$0.073 & 0.502$\pm$0.054 & 0.572$\pm$0.043 &0.474$\pm$0.057 & 0.447$\pm$0.052 & 0.547$\pm$0.059 &0.463$\pm$0.033 & 0.418$\pm$0.031 & 0.532$\pm$0.050 &{${4.9}^{***}$}\\
\ProGNN & O.O.T. & O.O.T. & O.O.T. & O.O.T. & O.O.T. & O.O.T. & O.O.T. & O.O.T. & O.O.T. & {N.A.}\\
\PTDNet & O.O.M. & O.O.M. & O.O.M. & O.O.M. & O.O.M. & O.O.M. & O.O.M. & O.O.M. & O.O.M. & {N.A.}\\
\FGC & 0.377$\pm$0.000 & 0.071$\pm$0.001 & 0.145$\pm$0.003 &0.366$\pm$0.000 & 0.055$\pm$0.000 & 0.103$\pm$0.001 &0.362$\pm$0.000 & 0.048$\pm$0.000 & 0.084$\pm$0.001 &{${9.6}^{***}$}\\
\hline
\hline
\Our & \colorbox{Red!40}{0.562$\pm$0.015} &  \colorbox{Red!40}{0.566$\pm$0.017} &  \colorbox{Green!40}{0.675$\pm$0.008} & \colorbox{Red!40}{0.528$\pm$0.020} &  \colorbox{Red!40}{0.520$\pm$0.013} &  \colorbox{Red!40}{0.664$\pm$0.007} & \colorbox{Red!40}{0.508$\pm$0.014} &  \colorbox{Red!40}{0.498$\pm$0.009} &  \colorbox{Red!40}{0.658$\pm$0.006} & {$\boldsymbol{1.2}$}\\
                \hline
            \end{tabular}
        }
        \vspace{0.15mm}
        \caption{Amazon-Photo}\label{tab:result:amazon-photo}
        \vspace{0.25mm}
    \end{subtable}
    \begin{subtable}{\textwidth}
        \centering
        \resizebox{\textwidth}{!}{
            \begin{tabular}{ c||c|c|c||c|c|c||c|c|c||c } 
                \hline
                 \textbf{Noise Level} & \multicolumn{3}{c||}{\rom{1}} & \multicolumn{3}{c||}{\rom{2}} & \multicolumn{3}{c||}{\rom{3}} & \multirow{2}{*}{\textbf{{AR}}} \\
                 \cline{1-10}
                 \textbf{Metric} & F1 Score &  NMI & Modularity & F1 Score &  NMI & Modularity & F1 Score &  NMI & Modularity & \\
                \hline
                \hline
\DeepWalk & 0.300$\pm$0.024 & 0.243$\pm$0.010 & \colorbox{Green!40}{0.680$\pm$0.009} & 0.216$\pm$0.010 & 0.155$\pm$0.006 & 0.593$\pm$0.011 &0.169$\pm$0.014 & 0.111$\pm$0.009 & 0.528$\pm$0.008 &{${8.3}^{***}$}\\
\ntov & 0.292$\pm$0.028 & 0.247$\pm$0.015 & \colorbox{Blue!40}{0.684$\pm$0.009} & 0.210$\pm$0.016 & 0.154$\pm$0.010 & 0.594$\pm$0.009 &0.170$\pm$0.009 & 0.111$\pm$0.011 & 0.528$\pm$0.009 &{${8.3}^{***}$}\\
\DGI & 0.351$\pm$0.040 & \colorbox{Red!40}{0.415$\pm$0.011} &  0.619$\pm$0.015 &0.294$\pm$0.027 & \colorbox{Red!40}{0.330$\pm$0.012} &  0.547$\pm$0.017 &0.248$\pm$0.018 & 0.240$\pm$0.017 & 0.412$\pm$0.033 &{${6.1}^{***}$}\\
\GMI & 0.277$\pm$0.023 & 0.319$\pm$0.008 & 0.576$\pm$0.010 &0.226$\pm$0.016 & 0.229$\pm$0.005 & 0.496$\pm$0.007 &0.152$\pm$0.016 & 0.145$\pm$0.012 & 0.391$\pm$0.020 &{${9.7}^{***}$}\\
\hline
\hline
\MinCutPool & 0.265$\pm$0.035 & 0.222$\pm$0.023 & 0.614$\pm$0.012 &0.217$\pm$0.027 & 0.147$\pm$0.019 & 0.556$\pm$0.012 &0.219$\pm$0.097 & 0.086$\pm$0.039 & 0.436$\pm$0.172 &{${10.1}^{***}$}\\
\DMoN & 0.400$\pm$0.023 & 0.343$\pm$0.015 & 0.661$\pm$0.012 &0.355$\pm$0.023 & 0.280$\pm$0.013 & \colorbox{Green!40}{0.620$\pm$0.013} & 0.326$\pm$0.016 & 0.231$\pm$0.016 & \colorbox{Green!40}{0.576$\pm$0.012} & {${4.6}^{***}$}\\
\GCC & 0.375$\pm$0.017 & 0.230$\pm$0.013 & 0.486$\pm$0.011 &0.364$\pm$0.023 & 0.114$\pm$0.014 & 0.312$\pm$0.053 &\colorbox{Red!40}{0.364$\pm$0.041} &  0.076$\pm$0.016 & 0.252$\pm$0.073 &{${9.6}^{***}$}\\
\hline
\hline
\Jaccard & \colorbox{Red!40}{0.415$\pm$0.022} &  \colorbox{Green!40}{0.364$\pm$0.017} &  0.661$\pm$0.014 &0.369$\pm$0.030 & \colorbox{Green!40}{0.310$\pm$0.014} &  \colorbox{Blue!40}{0.627$\pm$0.013} & \colorbox{Blue!40}{0.348$\pm$0.030} &  \colorbox{Blue!40}{0.276$\pm$0.017} &  \colorbox{Blue!40}{0.602$\pm$0.016} & {${2.7}^{**}$}\\
\SVD & 0.313$\pm$0.025 & 0.207$\pm$0.019 & 0.487$\pm$0.022 &0.291$\pm$0.031 & 0.172$\pm$0.023 & 0.468$\pm$0.016 &0.288$\pm$0.024 & 0.156$\pm$0.017 & 0.458$\pm$0.018 &{${8.9}^{***}$}\\
\GDC & 0.298$\pm$0.030 & 0.218$\pm$0.021 & 0.577$\pm$0.020 &0.266$\pm$0.027 & 0.183$\pm$0.017 & 0.555$\pm$0.011 &0.269$\pm$0.010 & 0.175$\pm$0.016 & 0.540$\pm$0.015 &{${8.1}^{***}$}\\
\ProGNN & \colorbox{Green!40}{0.405$\pm$0.023} &  0.348$\pm$0.015 & 0.631$\pm$0.015 &\colorbox{Green!40}{0.370$\pm$0.022} &  0.296$\pm$0.011 & 0.590$\pm$0.016 &0.341$\pm$0.018 & \colorbox{Green!40}{0.248$\pm$0.017} &  0.544$\pm$0.019 &{${4.2}^{***}$}\\
\PTDNet & 0.198$\pm$0.014 & 0.033$\pm$0.010 & 0.300$\pm$0.011 &0.186$\pm$0.010 & 0.031$\pm$0.005 & 0.279$\pm$0.007 &0.209$\pm$0.018 & 0.025$\pm$0.003 & 0.256$\pm$0.009 &{${13.6}^{***}$}\\
\FGC & 0.388$\pm$0.005 & 0.145$\pm$0.005 & 0.337$\pm$0.006 &\colorbox{Red!40}{0.374$\pm$0.005} &  0.123$\pm$0.006 & 0.314$\pm$0.006 &\colorbox{Red!40}{0.364$\pm$0.010} &  0.112$\pm$0.008 & 0.295$\pm$0.005 &{${8.7}^{***}$}\\
\hline
\hline
\Our & \colorbox{Blue!40}{0.413$\pm$0.030} &  \colorbox{Blue!40}{0.379$\pm$0.027} &  \colorbox{Red!40}{0.696$\pm$0.010} & \colorbox{Blue!40}{0.372$\pm$0.028} &  \colorbox{Blue!40}{0.320$\pm$0.023} &  \colorbox{Red!40}{0.660$\pm$0.015} & \colorbox{Blue!40}{0.348$\pm$0.028} &  \colorbox{Red!40}{0.282$\pm$0.021} &  \colorbox{Red!40}{0.628$\pm$0.018} & {$\boldsymbol{1.7}$}\\
                \hline
            \end{tabular}
        }
        \vspace{0.15mm}
        \caption{Cora}\label{tab:result:cora}
        \vspace{0.25mm}
    \end{subtable}
    \begin{subtable}{\textwidth}
        \vspace{1mm}
        \centering
        \resizebox{\textwidth}{!}{
            \begin{tabular}{ c||c|c|c||c|c|c||c|c|c||c } 
                \hline
                 \textbf{Noise Level} & \multicolumn{3}{c||}{\rom{1}} & \multicolumn{3}{c||}{\rom{2}} & \multicolumn{3}{c||}{\rom{3}} & \multirow{2}{*}{\textbf{{AR}}} \\
                 \cline{1-10}
                 \textbf{Metric} & F1 Score &  NMI & Modularity &F1 Score &  NMI & Modularity & F1 Score &  NMI & Modularity & \\
                \hline
                \hline                
\DeepWalk & 0.375$\pm$0.009 & 0.276$\pm$0.013 & 0.636$\pm$0.016 &0.314$\pm$0.010 & 0.201$\pm$0.012 & 0.581$\pm$0.009 &0.250$\pm$0.018 & 0.147$\pm$0.011 & 0.535$\pm$0.010 &{${6.4}^{***}$}\\
\ntov & 0.377$\pm$0.007 & 0.282$\pm$0.006 & 0.644$\pm$0.005 &0.308$\pm$0.010 & 0.199$\pm$0.011 & 0.584$\pm$0.006 &0.263$\pm$0.015 & 0.152$\pm$0.010 & 0.540$\pm$0.008 &{${5.8}^{***}$}\\
\DGI & 0.379$\pm$0.063 & 0.295$\pm$0.037 & 0.340$\pm$0.045 &0.242$\pm$0.009 & 0.131$\pm$0.015 & 0.167$\pm$0.021 &0.210$\pm$0.015 & 0.063$\pm$0.017 & 0.083$\pm$0.027 &{${9.8}^{***}$}\\
\GMI & 0.366$\pm$0.018 & 0.268$\pm$0.011 & 0.395$\pm$0.013 &0.259$\pm$0.012 & 0.144$\pm$0.007 & 0.234$\pm$0.017 &0.201$\pm$0.023 & 0.062$\pm$0.005 & 0.115$\pm$0.008 &{${10.1}^{***}$}\\
\hline
\hline
\MinCutPool & 0.271$\pm$0.026 & 0.200$\pm$0.019 & 0.592$\pm$0.018 &0.278$\pm$0.103 & 0.105$\pm$0.054 & 0.398$\pm$0.236 &\colorbox{Red!40}{0.437$\pm$0.067} &  0.012$\pm$0.026 & 0.035$\pm$0.124 &{${9.3}^{***}$}\\
\DMoN & 0.340$\pm$0.026 & 0.289$\pm$0.025 & \colorbox{Green!40}{0.661$\pm$0.016} & 0.314$\pm$0.020 & \colorbox{Green!40}{0.237$\pm$0.023} &  \colorbox{Green!40}{0.630$\pm$0.016} & 0.291$\pm$0.018 & \colorbox{Green!40}{0.204$\pm$0.019} &  \colorbox{Green!40}{0.600$\pm$0.016} & {${4.8}^{***}$}\\
\GCC & \colorbox{Red!40}{0.461$\pm$0.022} &  \colorbox{Blue!40}{0.299$\pm$0.024} &  0.441$\pm$0.041 &\colorbox{Red!40}{0.415$\pm$0.014} &  0.165$\pm$0.049 & 0.306$\pm$0.086 &\colorbox{Green!40}{0.379$\pm$0.030} &  0.105$\pm$0.036 & 0.232$\pm$0.063 &{${5.7}^{***}$}\\
\hline
\hline
\Jaccard & 0.358$\pm$0.033 & 0.283$\pm$0.033 & 0.600$\pm$0.015 &0.323$\pm$0.025 & 0.232$\pm$0.025 & 0.569$\pm$0.011 &0.295$\pm$0.020 & 0.200$\pm$0.019 & 0.538$\pm$0.017 &{${5.4}^{***}$}\\
\SVD & 0.275$\pm$0.022 & 0.165$\pm$0.024 & 0.403$\pm$0.029 &0.247$\pm$0.017 & 0.142$\pm$0.011 & 0.365$\pm$0.015 &0.261$\pm$0.016 & 0.129$\pm$0.011 & 0.338$\pm$0.016 &{${9.4}^{***}$}\\
\GDC & 0.267$\pm$0.019 & 0.159$\pm$0.016 & 0.475$\pm$0.019 &0.230$\pm$0.020 & 0.102$\pm$0.012 & 0.366$\pm$0.013 &0.190$\pm$0.024 & 0.060$\pm$0.014 & 0.285$\pm$0.009 &{${11.0}^{***}$}\\
\ProGNN & 0.345$\pm$0.026 & \colorbox{Green!40}{0.297$\pm$0.025} &  \colorbox{Blue!40}{0.662$\pm$0.016} & 0.319$\pm$0.020 & \colorbox{Blue!40}{0.244$\pm$0.023} &  \colorbox{Blue!40}{0.631$\pm$0.015} & 0.297$\pm$0.018 & \colorbox{Blue!40}{0.212$\pm$0.020} &  \colorbox{Blue!40}{0.603$\pm$0.015} & {${3.6}^{***}$}\\
\PTDNet & 0.235$\pm$0.045 & 0.058$\pm$0.007 & 0.288$\pm$0.007 &0.216$\pm$0.055 & 0.046$\pm$0.006 & 0.249$\pm$0.012 &0.271$\pm$0.041 & 0.044$\pm$0.006 & 0.244$\pm$0.013 &{${11.9}^{***}$}\\
\FGC & \colorbox{Blue!40}{0.395$\pm$0.013} &  0.035$\pm$0.011 & 0.111$\pm$0.029 &\colorbox{Red!40}{0.415$\pm$0.002} &  0.017$\pm$0.003 & 0.059$\pm$0.005 &\colorbox{Blue!40}{0.424$\pm$0.003} &  0.010$\pm$0.001 & 0.038$\pm$0.006 &{${9.8}^{***}$}\\
\hline
\hline
\Our & \colorbox{Green!40}{0.380$\pm$0.034} &  \colorbox{Red!40}{0.337$\pm$0.024} &  \colorbox{Red!40}{0.683$\pm$0.014} & \colorbox{Blue!40}{0.333$\pm$0.026} &  \colorbox{Red!40}{0.282$\pm$0.015} &  \colorbox{Red!40}{0.656$\pm$0.015} & 0.319$\pm$0.021 & \colorbox{Red!40}{0.255$\pm$0.014} &  \colorbox{Red!40}{0.639$\pm$0.015} & {$\boldsymbol{1.8}$}\\
                \hline
            \end{tabular}
        }
        \vspace{0.15mm}
        \caption{Cora-ML}\label{tab:result:cora_ml} 
        \vspace{0.25mm}
    \end{subtable}
    \raggedright
    \textbf{(continues on the next page)} \\
\end{table*}

\begin{table*}[h!]
\vspace{1mm}
\end{table*}

\begin{table*}[h!]
    \vspace{-4mm}
    \ContinuedFloat
    \centering
    \begin{subtable}{\textwidth}
        \vspace{1mm}
        \centering
        \resizebox{\textwidth}{!}{
            \begin{tabular}{ c||c|c|c||c|c|c||c|c|c||c } 
                \hline
                 \textbf{Noise Level} & \multicolumn{3}{c||}{\rom{1}} & \multicolumn{3}{c||}{\rom{2}} & \multicolumn{3}{c||}{\rom{3}} & \multirow{2}{*}{\textbf{{AR}}} \\
                 \cline{1-10}
                 \textbf{Metric} & F1 Score &  NMI & Modularity &F1 Score &  NMI & Modularity & F1 Score &  NMI & Modularity & \\
                \hline
                \hline
\DeepWalk & 0.128$\pm$0.004 & 0.089$\pm$0.003 & \colorbox{Red!40}{0.650$\pm$0.004} & 0.103$\pm$0.004 & 0.053$\pm$0.003 & \colorbox{Blue!40}{0.586$\pm$0.005} & 0.086$\pm$0.004 & 0.037$\pm$0.002 & \colorbox{Red!40}{0.545$\pm$0.003} & {${7.3}^{***}$}\\
\ntov & 0.127$\pm$0.004 & 0.089$\pm$0.003 & \colorbox{Red!40}{0.650$\pm$0.003} & 0.101$\pm$0.005 & 0.053$\pm$0.002 & \colorbox{Red!40}{0.587$\pm$0.005} & 0.085$\pm$0.004 & 0.037$\pm$0.003 & \colorbox{Red!40}{0.545$\pm$0.005} & {${7.6}^{***}$}\\
\DGI & 0.199$\pm$0.019 & 0.108$\pm$0.003 & 0.169$\pm$0.005 &0.160$\pm$0.007 & 0.064$\pm$0.002 & 0.153$\pm$0.007 &0.130$\pm$0.006 & 0.043$\pm$0.003 & 0.160$\pm$0.007 &{${9.2}^{***}$}\\
\GMI & 0.159$\pm$0.009 & 0.117$\pm$0.002 & 0.154$\pm$0.003 &0.115$\pm$0.010 & 0.072$\pm$0.001 & 0.135$\pm$0.002 &0.116$\pm$0.006 & 0.049$\pm$0.001 & 0.115$\pm$0.006 &{${9.3}^{***}$}\\
\hline
\hline
\MinCutPool & 0.380$\pm$0.005 & 0.131$\pm$0.005 & 0.512$\pm$0.010 &0.345$\pm$0.017 & 0.097$\pm$0.009 & 0.487$\pm$0.011 &0.325$\pm$0.020 & 0.079$\pm$0.011 & 0.472$\pm$0.016 &{${5.9}^{***}$}\\
\DMoN & 0.406$\pm$0.005 & \colorbox{Green!40}{0.161$\pm$0.004} &  \colorbox{Blue!40}{0.542$\pm$0.001} & 0.377$\pm$0.008 & \colorbox{Green!40}{0.125$\pm$0.011} &  0.518$\pm$0.003 &0.346$\pm$0.019 & \colorbox{Blue!40}{0.090$\pm$0.023} &  0.497$\pm$0.010 &{${3.7}^{***}$}\\
\GCC & \colorbox{Blue!40}{0.522$\pm$0.002} &  0.052$\pm$0.003 & 0.313$\pm$0.005 &\colorbox{Blue!40}{0.459$\pm$0.004} &  0.039$\pm$0.001 & 0.421$\pm$0.004 &\colorbox{Blue!40}{0.505$\pm$0.053} &  0.019$\pm$0.010 & 0.276$\pm$0.149 &{${7.4}^{***}$}\\
\hline
\hline
\Jaccard & 0.407$\pm$0.005 & \colorbox{Blue!40}{0.163$\pm$0.005} &  \colorbox{Blue!40}{0.542$\pm$0.001} & 0.377$\pm$0.008 & \colorbox{Blue!40}{0.126$\pm$0.011} &  0.518$\pm$0.003 &0.346$\pm$0.020 & \colorbox{Blue!40}{0.090$\pm$0.023} &  0.498$\pm$0.010 &{${3.2}^{***}$}\\
\SVD & 0.372$\pm$0.043 & 0.094$\pm$0.022 & 0.379$\pm$0.006 &0.351$\pm$0.038 & 0.076$\pm$0.014 & 0.372$\pm$0.006 &0.332$\pm$0.035 & 0.060$\pm$0.014 & 0.365$\pm$0.005 &{${7.1}^{***}$}\\
\GDC & 0.360$\pm$0.022 & 0.113$\pm$0.013 & 0.481$\pm$0.010 &0.351$\pm$0.006 & 0.097$\pm$0.005 & 0.477$\pm$0.004 &0.327$\pm$0.023 & 0.071$\pm$0.022 & 0.475$\pm$0.011 &{${6.2}^{***}$}\\
\ProGNN & O.O.T. & O.O.T. & O.O.T. & O.O.T. & O.O.T. & O.O.T. & O.O.T. & O.O.T. & O.O.T. & {N.A.}\\
\PTDNet & O.O.M. & O.O.M. & O.O.M. & O.O.M. & O.O.M. & O.O.M. & O.O.M. & O.O.M. & O.O.M. & {N.A.}\\
\FGC & \colorbox{Red!40}{0.598$\pm$0.000} &  0.000$\pm$0.000 & 0.000$\pm$0.000 &\colorbox{Red!40}{0.568$\pm$0.000} &  0.059$\pm$0.000 & 0.261$\pm$0.000 &\colorbox{Red!40}{0.576$\pm$0.000} &  0.044$\pm$0.000 & 0.218$\pm$0.000 &{${7.1}^{***}$}\\
\hline
\hline
\Our & \colorbox{Green!40}{0.414$\pm$0.009} &  \colorbox{Red!40}{0.175$\pm$0.010} &  0.540$\pm$0.001 &\colorbox{Green!40}{0.396$\pm$0.004} &  \colorbox{Red!40}{0.160$\pm$0.004} &  \colorbox{Green!40}{0.523$\pm$0.001} & \colorbox{Green!40}{0.380$\pm$0.003} &  \colorbox{Red!40}{0.141$\pm$0.004} &  \colorbox{Blue!40}{0.513$\pm$0.001} & {$\boldsymbol{2.6}$}\\
                \hline
            \end{tabular}
        }
        \vspace{0.15mm}
        \caption{Pubmed}\label{tab:result:pubmed}
        \vspace{0.25mm}
    \end{subtable}
    \begin{subtable}{\textwidth}
        \vspace{1mm}
        \centering
        \resizebox{\textwidth}{!}{
            \begin{tabular}{ c||c|c|c||c|c|c||c|c|c||c } 
                \hline
                 \textbf{Noise Level} & \multicolumn{3}{c||}{\rom{1}} & \multicolumn{3}{c||}{\rom{2}} & \multicolumn{3}{c||}{\rom{3}} & \multirow{2}{*}{\textbf{{AR}}} \\
                 \cline{1-10}
                 \textbf{Metric} & F1 Score &  NMI & Modularity &F1 Score &  NMI & Modularity & F1 Score &  NMI & Modularity & \\
                \hline
                \hline  
\DeepWalk & 0.177$\pm$0.013 & 0.083$\pm$0.005 & \colorbox{Red!40}{0.741$\pm$0.006} & 0.136$\pm$0.012 & 0.054$\pm$0.007 & \colorbox{Green!40}{0.656$\pm$0.010} & 0.112$\pm$0.014 & 0.044$\pm$0.004 & \colorbox{Green!40}{0.596$\pm$0.007} & {${9.7}^{***}$}\\
\ntov & 0.180$\pm$0.011 & 0.084$\pm$0.005 & \colorbox{Blue!40}{0.740$\pm$0.006} & 0.146$\pm$0.013 & 0.057$\pm$0.004 & \colorbox{Blue!40}{0.661$\pm$0.007} & 0.116$\pm$0.011 & 0.041$\pm$0.005 & \colorbox{Green!40}{0.596$\pm$0.011} & {${9.2}^{***}$}\\
\DGI & 0.256$\pm$0.021 & \colorbox{Red!40}{0.294$\pm$0.006} &  0.704$\pm$0.015 &0.204$\pm$0.013 & \colorbox{Blue!40}{0.228$\pm$0.006} &  0.648$\pm$0.020 &0.183$\pm$0.020 & \colorbox{Blue!40}{0.176$\pm$0.004} &  0.562$\pm$0.032 &{${6.0}^{***}$}\\
\GMI & 0.248$\pm$0.016 & \colorbox{Blue!40}{0.292$\pm$0.006} &  0.606$\pm$0.011 &0.210$\pm$0.011 & \colorbox{Red!40}{0.238$\pm$0.004} &  0.547$\pm$0.008 &0.185$\pm$0.009 & \colorbox{Red!40}{0.199$\pm$0.006} &  0.486$\pm$0.015 &{${7.1}^{***}$}\\
\hline
\hline
\MinCutPool & 0.267$\pm$0.034 & 0.157$\pm$0.024 & 0.677$\pm$0.012 &\colorbox{Green!40}{0.350$\pm$0.130} &  0.067$\pm$0.050 & 0.385$\pm$0.287 &\colorbox{Red!40}{0.435$\pm$0.136} &  0.021$\pm$0.037 & 0.154$\pm$0.256 &{${8.6}^{***}$}\\
\DMoN & 0.346$\pm$0.024 & 0.182$\pm$0.017 & 0.665$\pm$0.011 &0.308$\pm$0.012 & 0.139$\pm$0.009 & 0.624$\pm$0.008 &0.283$\pm$0.009 & 0.112$\pm$0.006 & 0.591$\pm$0.009 &{${6.6}^{***}$}\\
\GCC & \colorbox{Red!40}{0.410$\pm$0.011} &  0.191$\pm$0.031 & 0.545$\pm$0.066 &\colorbox{Red!40}{0.415$\pm$0.024} &  0.147$\pm$0.039 & 0.448$\pm$0.114 &\colorbox{Blue!40}{0.419$\pm$0.054} &  0.078$\pm$0.039 & 0.312$\pm$0.161 &{${6.3}^{***}$}\\
\hline
\hline
\Jaccard & \colorbox{Blue!40}{0.369$\pm$0.031} &  0.209$\pm$0.018 & 0.676$\pm$0.009 &0.337$\pm$0.011 & 0.171$\pm$0.008 & 0.643$\pm$0.004 &0.306$\pm$0.012 & 0.139$\pm$0.007 & \colorbox{Blue!40}{0.612$\pm$0.007} & {${4.0}^{***}$}\\
\SVD & 0.280$\pm$0.033 & 0.120$\pm$0.013 & 0.448$\pm$0.021 &0.248$\pm$0.027 & 0.084$\pm$0.008 & 0.422$\pm$0.022 &0.237$\pm$0.027 & 0.062$\pm$0.010 & 0.398$\pm$0.022 &{${9.8}^{***}$}\\
\GDC & 0.257$\pm$0.026 & 0.117$\pm$0.015 & 0.548$\pm$0.021 &0.231$\pm$0.015 & 0.096$\pm$0.012 & 0.530$\pm$0.013 &0.232$\pm$0.018 & 0.089$\pm$0.012 & 0.529$\pm$0.014 &{${9.4}^{***}$}\\
\ProGNN & 0.359$\pm$0.025 & 0.191$\pm$0.017 & 0.636$\pm$0.012 &0.326$\pm$0.016 & 0.153$\pm$0.009 & 0.587$\pm$0.013 &0.302$\pm$0.012 & 0.125$\pm$0.006 & 0.544$\pm$0.012 &{${5.9}^{***}$}\\
\PTDNet & 0.278$\pm$0.029 & 0.048$\pm$0.004 & 0.344$\pm$0.007 &0.277$\pm$0.044 & 0.036$\pm$0.014 & 0.317$\pm$0.019 &0.293$\pm$0.037 & 0.056$\pm$0.025 & 0.301$\pm$0.018 &{${11.3}^{***}$}\\
\FGC & \colorbox{Red!40}{0.410$\pm$0.004} &  0.131$\pm$0.005 & 0.409$\pm$0.007 &\colorbox{Blue!40}{0.398$\pm$0.005} &  0.112$\pm$0.007 & 0.381$\pm$0.008 &\colorbox{Green!40}{0.400$\pm$0.005} &  0.105$\pm$0.005 & 0.370$\pm$0.008 &{${7.4}^{***}$}\\
\hline
\hline
\Our & 0.363$\pm$0.017 & \colorbox{Green!40}{0.230$\pm$0.013} &  \colorbox{Green!40}{0.707$\pm$0.007} & 0.330$\pm$0.025 & \colorbox{Green!40}{0.194$\pm$0.021} &  \colorbox{Red!40}{0.677$\pm$0.012} & 0.289$\pm$0.017 & \colorbox{Green!40}{0.151$\pm$0.013} &  \colorbox{Red!40}{0.640$\pm$0.009} & {$\boldsymbol{3.3}$}\\
                \hline
            \end{tabular}
        }
        \vspace{0.15mm}
        \caption{Citeseer}\label{tab:result:citeseer}
        \vspace{0.25mm}
    \end{subtable}
\end{table*}

\subsection{Q3. Ablation Study}
\label{sec:exp:ablation}

We have just shown that the meta-model in \Our is effective in distinguishing real edges and noise edges.
We further examine how much it affects the performance of \Our by comparing \Our with two variants of it:
\begin{itemize}[leftmargin=*]
    \item \textbf{\OurAblOne}: \Our without the meta-model, i.e., the weight $V_{ij}$ of every pair $(v_i, v_j)$ is the same.
    \item \textbf{\OurAblTwo}: \Our with the metal model using only node attributes, spec., $(Y^{(1)}_{ij}, Y^{(2)}_{ij}, Y^{(3)}_{ij}) = (1, 0, 0)$ for every pair $(v_i, v_j)$.
\end{itemize}
In Table~\ref{tab:ablation}, we report the results of the ablation study on Citeseer.
For each noise level, each metric, and each variant including the original \Our, we report the mean and standard deviation of the results of the 15 trials.
In all the settings and in all respects, the original \Our performs best, and \OurAblOne without the meta-model performs worst, validating the performance boost of the meta-model in \Our.
Moreover, the comparison between \Our and \OurAblTwo shows that using the additional information on the topology-based node similarity and the soft cluster assignment matrix from the clustering model is helpful. 
Interestingly, meta-weighting is more helpful w.r.t F1 Score and NMI than w.r.t modularity, although the meta-model is updated using the modularity-based objective.

\begin{table*}[t!]
    \vspace{-3mm}
    \centering
    \caption{\label{tab:effective} (Q2) \underline{\smash{Effectiveness of Meta-Weighting.}}
    {\Our successfully assigns high weights to real edges and low weights to noise edges.
    The PRAUC and HITS@10\% values are consistently and significantly higher than the baseline. The baseline is the expected value of PRAUC and HITS@10\% when weights are randomly assigned.}
    }
    \scalebox{0.9}{
        \begin{tabular}{ c||c|c|c||c|c|c||c|c|c||c|c|c||c|c|c } 
            \hline
             \textbf{Dataset} & \multicolumn{3}{c||}{\textbf{Cora}} & \multicolumn{3}{c||}{\textbf{Cora-ML}} &\multicolumn{3}{c||}{\textbf{Citeseer}} & \multicolumn{3}{c||}{\textbf{Amazon-Photo}} & \multicolumn{3}{c}{\textbf{Pubmed}} \\
             \hline
             \textbf{Noise Level} & \rom{1} &  \rom{2} & \rom{3} & \rom{1} &  \rom{2} & \rom{3} & \rom{1} &  \rom{2} & \rom{3} & \rom{1} &  \rom{2} & \rom{3} & \rom{1} &  \rom{2} & \rom{3} \\
            \hline
            \hline
            PRAUC & 0.927 & 0.875 & 0.831 & 0.934 & 0.878 & 0.825  & 0.908 & 0.843 & 0.793 & 0.993 & 0.985 & 0.976 & 0.890 & 0.813 & 0.757 \\
            HITS@10\% & 0.999 & 0.997 & 0.993 & 1.000 & 0.997 & 0.988  & 0.999 & 0.995 & 0.991 & 1.00 & 0.999 & 0.999 & 0.999 & 0.997 & 0.991 \\
            \hline
             Baseline & 0.769 & 0.625 & 0.526 & 0.769 & 0.625 & 0.526  & 0.769 & 0.625 & 0.526 & 0.769 & 0.625 & 0.526 & 0.769 & 0.625 & 0.526 \\
            \hline
        \end{tabular}
    }
\end{table*}

\begin{figure*}[t!]
    \vspace{-1mm}
    \includegraphics[width=0.4\linewidth]{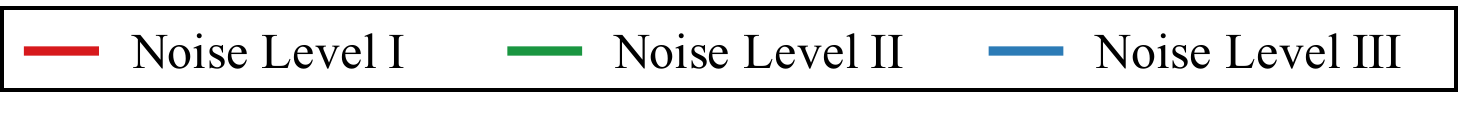} \\
    \begin{subfigure}{0.195\linewidth}
        \centering
        \includegraphics[width=\linewidth]{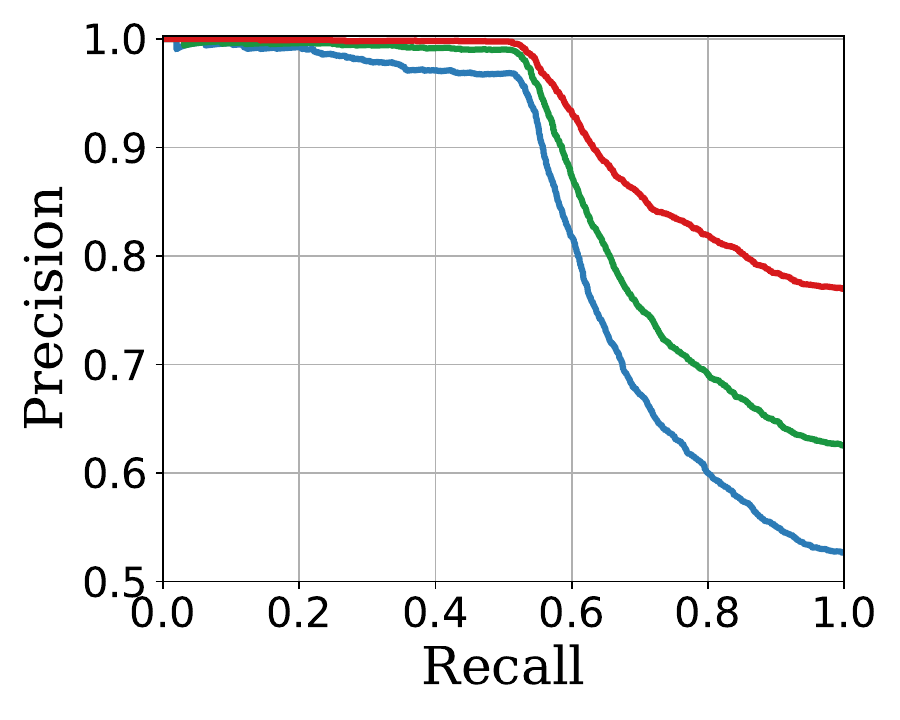}
        \caption{Cora}
        \label{fig:prauc:cora}
    \end{subfigure}
    \hfill
    \begin{subfigure}{0.195\linewidth}
        \centering
        \includegraphics[width=\textwidth]{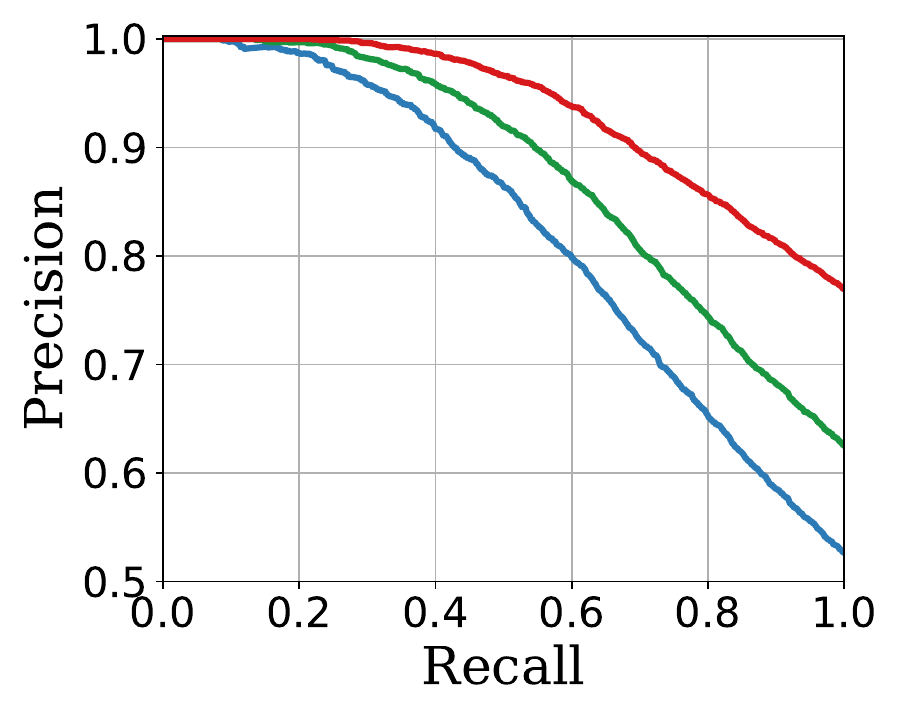}
        \caption{Cora-ML}
        \label{fig:prauc:cora_ml}
    \end{subfigure}
    \hfill
    \begin{subfigure}{0.195\linewidth}
        \centering
        \includegraphics[width=\linewidth]{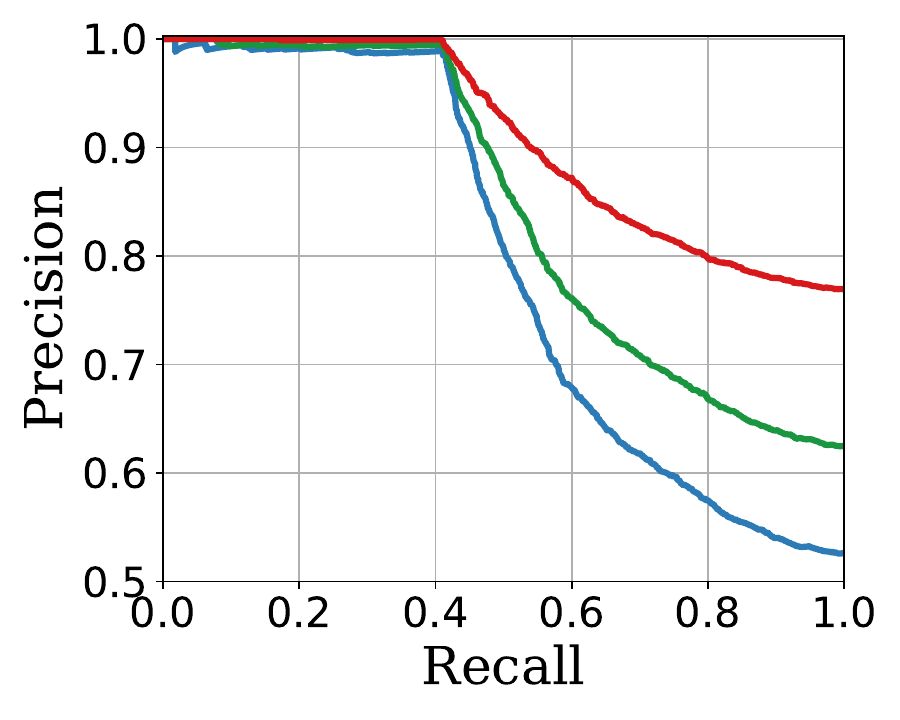}
        \caption{Citeseer}
        \label{fig:prauc:citeseer}
    \end{subfigure} 
    \hfill
    \begin{subfigure}{0.195\linewidth}
        \centering
        \includegraphics[width=\linewidth]{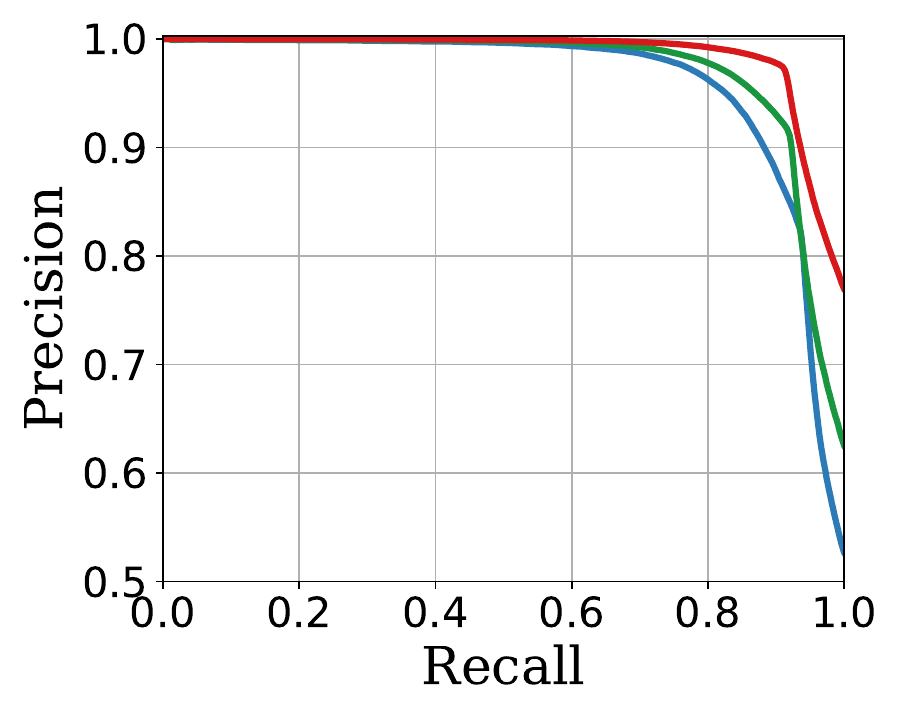}
        \caption{Amazon-Photo}
        \label{fig:prauc:amazon_photo}
    \end{subfigure} 
    \hfill
    \begin{subfigure}{0.195\linewidth}
        \centering
        \includegraphics[width=\linewidth]{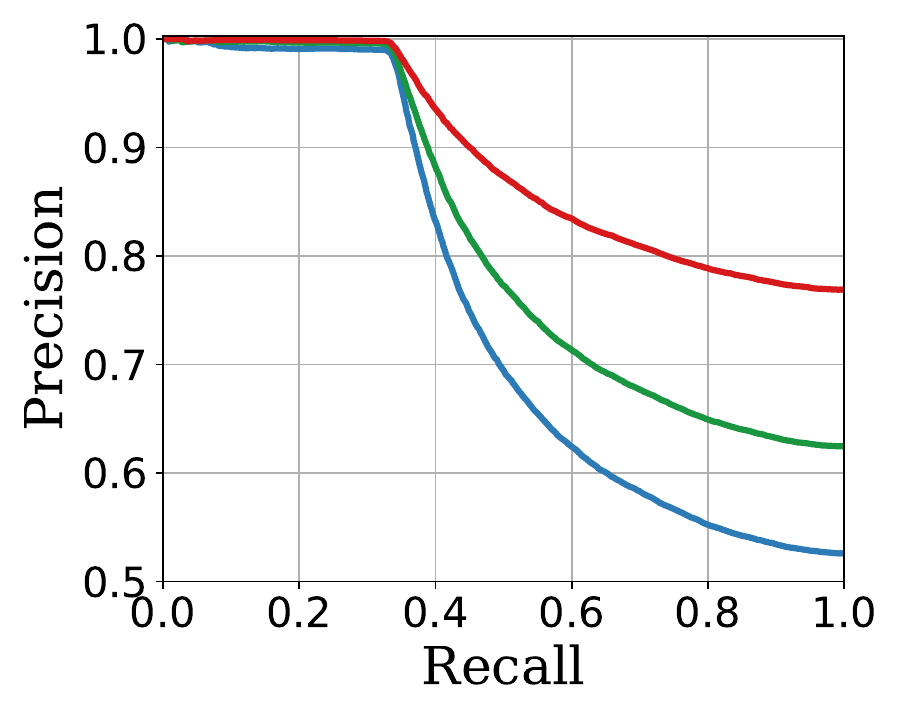}
        \caption{Pubmed}
        \label{fig:prauc:pubmed}
    \end{subfigure} 
    \caption{Detailed Precision-Recall curves used for Table~\ref{tab:effective}.}
    \label{fig:prauc}
\end{figure*}

\begin{table*}[t!]
    \vspace{-2mm}
    \centering
    \caption{\label{tab:ablation} (Q3) \underline{\smash{Ablation Study.}}
    Meta-weighting in \Our improves the performance (compare \Our and \OurAblOne).
    Using the topology-based node similarity and the output of the clustering model is also helpful (compare \Our and \OurAblTwo).
    }
    \scalebox{0.87}{
        \begin{tabular}{ c||c|c|c||c|c|c||c|c|c } 
            \hline
             \textbf{Noise Level} & \multicolumn{3}{c||}{\rom{1}} & \multicolumn{3}{c||}{\rom{2}} & \multicolumn{3}{c}{\rom{3}} \\
             \cline{1-10}
             \textbf{Metric} & F1 Score &  NMI & Modularity &F1 Score &  NMI & Modularity & F1 Score &  NMI & Modularity \\
            \hline
            \hline
\OurAblOne & 0.340$\pm$0.022 & 0.203$\pm$0.016 & 0.695$\pm$0.006 & 0.308$\pm$0.017 & 0.173$\pm$0.014 & 0.662$\pm$0.007 & 0.280$\pm$0.018 & 0.142$\pm$0.013 & 0.634$\pm$0.009\\
\OurAblTwo & \underline{0.346$\pm$0.020} & \underline{0.214$\pm$0.014} & \underline{0.701$\pm$0.007} & \underline{0.324$\pm$0.019} & \underline{0.187$\pm$0.016} & \underline{0.674$\pm$0.011} & \underline{0.288$\pm$0.017} & \underline{0.150$\pm$0.012} & \underline{0.638$\pm$0.009}\\
\hline
\Our & \textbf{0.363$\pm$0.017} & \textbf{0.230$\pm$0.013} & \textbf{0.707$\pm$0.007} & \textbf{0.330$\pm$0.025} & \textbf{0.194$\pm$0.021} & \textbf{0.677$\pm$0.012} & \textbf{0.289$\pm$0.017} & \textbf{0.151$\pm$0.013} & \textbf{0.640$\pm$0.009}\\
            \hline
        \end{tabular}
    }
\end{table*}

\section{Conclusion}
\label{sec:conclusion}
In this work, we propose \Our for robust GNN-based graph clustering against noise edges.
\Our consists of a GNN-based clustering model using a decomposable loss function 
with theoretical justification,
and a meta-model that adaptively adjusts the weights of node pairs in the loss function.
In our extensive experiments on the five datasets under three levels of noise, 
\Our is robust against noise edges, achieving an average rank of $1.2$ to $3.3$ among all the $14$ considered methods.
We also demonstrate the effectiveness of the meta-model by showing that it
(a) assigns high weights to real edges and low weights to noise edges and 
(b) leads to a performance boost, especially when it uses richer information. 

\vspace{1mm}
\smallsection{Acknowledgements:} This work was supported by Institute of Information \&
Communications Technology Planning \& Evaluation (IITP) grant funded by the Korea government (MSIT) (No. 2022-0-00871, Development of AI Autonomy and Knowledge Enhancement for AI Agent Collaboration) (No. 2022-0-00157, Robust, Fair, Extensible Data-Centric Continual Learning) (No. 2019-0-00075, Artificial Intelligence Graduate School Program (KAIST)).

\appendix

\section{Appendix: Parameter Settings}
\label{appendix:param_settings}
In this section, we provide detailed parameter settings.
For a fair comparison, we set the embedding dimension of all the considered methods (including \Ours) to 64, unless otherwise stated.

\smallsection{\textbf{\Our}}: 
\Our consists of a meta-model and a clustering model.
The clustering model consists of a single-layer GCN with skip connections (see Sec.~\ref{sec:method:meta} of the main paper) and a single-layer perceptron. The number of hidden units is $64$ in the GCN, and the single-layer perceptron 
outputs the final cluster assignment vector, whose length is equal to the number of clusters.
In the meta-model, for each $q$, $H^{(q)}$ is obtained by a two-layer perceptron.
We use ReLU activation~\cite{nair2010rectified} after the first layer, and we do not use any activation after the second layer.
We train \Our with the Adam optimizer~\cite{kingma2014adam}. 
The learning rates of the clustering model and the meta-model are fine-tuned via a grid search, where the range of both learning rates is 
$\{5e-4, 1e-3, 2e-3, 3e-3, 4e-3, 5e-3\}$.
Similarly, the batch sizes of the two models are fine-tuned via a grid search, where the range of both batch sizes is $\{128, 256, 512, 1024, 2048\}$.
\Our is trained with at least $200$ epochs and at most $1500$ epochs. 
We terminate the training if the modularity is not improved in the last $50$ epochs, and use the parameters giving the highest modularity during the training.


\smallsection{\textbf{Node embedding-based methods}}:
For node-embedding-based methods, \DGI, \GMI, \DeepWalk, and \ntov, we cluster
their output embeddings using K-means++~\cite{vassilvitskii2006k}, for which
we set the maximum number of iterations to $300$ and the tolerance to $1e-4$.
For \DGI and \GMI, we increase the embedding dimension to $512$,\footnote{The performance of \DGI and \GMI degrades significantly if we set the embedding dimension as $64$.} as in the original papers; and we 
use the hyperparameters in the source code released by the authors.
For \DeepWalk and \ntov, we set the number of walks as $80$, the length of the walks $80$, and the window size $10$.

\smallsection{\textbf{GNN-based graph clustering methods}}:
We compare \Ours with three GNN-based graph clustering methods: \MinCutPool, \DMoN and \GCC.
\MinCutPool and \DMoN consist of a single-layer GCN with skip connections and a single-layer perceptron, as \Our does.
We use ELU activation~\cite{clevert2015fast} in \MinCutPool and SELU activation in \DMoN (as in \Ours), following the original papers.
Moreover, following the original paper, in \DMoN, we use a dropout layer before the softmax operation. 
The dropout ratio is set as 0.5.
\MinCutPool and \DMoN are trained with at least $2000$ epochs and at most $4000$ epochs. 
We terminate the training if the modularity is not improved in the last $100$ epochs, and use the parameters giving the highest modularity during the training.
For \GCC, we set the maximum number of optimization iterations as $30$ and the tolerance $1e-7$.
The propagation order is fine-tuned exhaustively in the range from $1$ to $150$.

\smallsection{\textbf{Graph denoising methods}}:
We compare \Ours with four graph denoising methods:
\Jaccard, \SVD, \GDC, and \ProGNN.
\Jaccard removes each edge such that the Jaccard similarity between the attributes of the two endpoints is $0$.
\SVD generates a rank-$100$ approximation of the adjacency matrix, and uses such an approximation instead of the original matrix.
In \GDC, we use personalized PageRank~\cite{page1999pagerank}, which performs best in the original paper.
For \ProGNN and \PTDNet (see Sec.~\ref{sec:prelim_related:robust} for more details),
we replace their GNN models with the clustering model employed by \Ours and replace their classification loss with our clustering loss (see Eq.~\eqref{eq:modularity_loss}).
The hyperparameters of \ProGNN and \PTDNet are fine-tuned within the range specified in the source code provided by the authors.
For \SVD and \GDC, which generate weighted graphs, we observe that
applying \DMoN directly on weighted graphs impairs the performance of graph clustering.
Therefore, we convert each generated weighted graph to an unweighted one consisting only of the $|E|$ edges with the highest weights, where $|E|$ is the number of edges in the original noise-free graph.
For \FGC, the node similarity matrix is optimized using the loss function in the original paper.
The order of the graph Laplacian filter is fine-tuned exhaustively in the range from $1$ to $15$,
and the trade-off parameter $\alpha$ in the range $\{0.0001, 0.01, 1, 10, 100\}$.


\bibliographystyle{ACM-Reference-Format}
\balance
\bibliography{ref.bib}
	
\end{document}